\documentclass[twoside,11pt]{article}

\usepackage{blindtext}

 \usepackage[abbrvbib, preprint]{jmlr2e}

\usepackage{tikz}
\usepackage{verbatim}
\usepackage{amsmath}
\usepackage{amssymb}
\usepackage{graphicx}
\usepackage{dsfont}
\usepackage{algorithm}
\usepackage{xcolor}
\usepackage{setspace}
\usepackage{appendix}
\usepackage{comment}
\usepackage{multirow}
\usepackage{caption}
\usepackage{subcaption}
\usepackage{wasysym}
\usepackage{makecell}
\usepackage{afterpage}
\usepackage{natbib}

\usepackage{array}
\newcolumntype{C}[1]{>{\centering\arraybackslash$}p{#1}<{$}}

\newcommand{\bs}{\boldsymbol}
\makeatletter
\newcommand{\vast}{\bBigg@{4}}
\newcommand{\Vast}{\bBigg@{5}}
\makeatother

\newcommand{\JG}[1]{{\color{magenta}JG: #1}}

\allowdisplaybreaks

\newtheorem{property}{Property}

\DeclareMathOperator*{\argmax}{argmax}

\usepackage{lastpage}

\ShortHeadings{Learning to Rank under Multinomial Logit Choice}{Grant and Leslie}
\firstpageno{1}

\begin{document}
\title{Learning to Rank under Multinomial Logit Choice}
\author{\name James A. Grant \email j.grant@lancaster.ac.uk  \\
               \addr Department of Mathematics and Statistics, \\ 
		Lancaster University, \\
		Lancaster, UK
		\AND
		\name David S. Leslie \email d.leslie@lancaster.ac.uk \\
		\addr Department of Mathematics and Statistics, \\ 
		Lancaster University, \\
		Lancaster, UK}

\maketitle

\begin{abstract}%
Learning the optimal ordering of content is an important challenge in website design. The learning to rank (LTR) framework models this problem as a sequential problem of selecting lists of content and observing where users decide to click. Most previous work on LTR assumes that the user considers each item in the list in isolation, and makes binary choices to click or not on each. We introduce a multinomial logit (MNL) choice model to the LTR framework, which captures the behaviour of users who consider the ordered list of items as a whole and make a single choice among all the items and a no-click option. Under the MNL model, the user favours items which are either inherently more attractive, or placed in a preferable position within the list. We propose upper confidence bound (UCB) algorithms to minimise regret in two settings - where the position dependent parameters are known, and unknown. We present theoretical analysis leading to an $\Omega(\sqrt{JT})$ lower bound for the problem, an $\tilde{O}(\sqrt{JT})$ upper bound on regret of the UCB algorithm in the known-parameter setting, and an $\tilde{O}(K^2\sqrt{JT})$ upper bound on regret, the first, in the more challenging unknown-position-parameter setting. Our analyses are based on tight new concentration results for Geometric random variables, and novel functional inequalities for maximum likelihood estimators computed on discrete data.\end{abstract}


\section{Introduction}
Learning the optimal ordering of content is an important challenge in website design and online advertising. The learning to rank (LTR) framework captures such a challenge via a sequential decision-making model. In this setting, a decision-maker repeatedly selects orderings of items (product advertisements, search results, news articles etc.) and displays them to a user visiting their website. In response the user opts to click on none, one, or more of the displayed items. The objective of the decision-maker will be to maximise the number of clicks received over many iterations of this process. Such an objective is a reasonable and widely-used proxy for the most common interests of a decision-maker in this setting: e.g. maximising profit, and maximising user satisfaction. As such, methods which achieve this objective can be hugely impactful in real-world settings.

Recent works (e.g. \cite{KvetonEtAl2015a}, \cite{LagreeEtAl2016}) have considered various formulations of LTR, distinguished by the assumptions on the click model, assumed to govern how users decide to click on items. A majority of previous works utilise a \emph{factored} click model, which assumes, in particular, that the user will click on any displayed item satisfying two conditions: 1) that the user finds the item \emph{attractive}, and 2) the user \emph{examines} the item. The various factored models are differentiated by their specification of the probability of attraction and examination events given particular orderings of content. 

Factored models represent the user's choice as a series of binary decisions to click or not click on each examined item. They fail to capture settings where the user's decisions are made among more than two alternatives, for instance choosing between several items considered simultaneously. In this work, we consider LTR under a click model which captures the phenomenon of a user making a \emph{single} decision among \emph{several} alternatives including the option to not click at all. Our click model is based on the \emph{multinomial logit} (MNL) model of discrete user choice \citep{Luce1959,Plackett1975}. We augment the classical MNL model with position effects to capture the relative prominence of different display positions, which may be pre-specified or learned online. The classic model is a special case of ours.

Our model may be more suitable in a variety of settings. For instance, where all items are visible to a user simultaneously 
or are laid out on a grid 
such that it is not possible, a priori, to specify a rank order of positions in terms of prominence, or assume that users consider items sequentially.  Several studies in the information retrieval and recommender systems literatures highlight the complexity of position effects in grid displays \citep{xie2017investigating,xie2019grid,guo2020debiasing}, and the issues with assuming items are considered independently by default \citep{oosterhuis2019optimizing,ccapan2022dirichlet,hazrati2022recommender}.


\subsection{Problem Definition}

We propose the \emph{Multinomial Logit Learning to Rank} (MNL-LTR) problem. This problem captures the challenge of learning an optimal list of $K$ items among $J$, where the click model is an order-dependent variant of multinomial logit choice. 

In each of a series of rounds $t \in [T]$,\footnote{For an integer $W \geq 1$, we let $[W]$ denote the set $\{1,\dots,W\}$} the decision-maker chooses an \emph{action} $\mathbf{a}_t = (a_{1,t},\dots,a_{K,t})\in \mathcal{A} \subset [J]^K$, where $\mathcal{A}$ is the set of all ordered lists of length $K$ consisting of items drawn from $[J]$ without replacement. The action indicates an ordering of $K$ items to display to the user in round $t$. Each action $j \in [J]$ has an associated \emph{attractiveness parameter}, $\alpha_j \in (0,1]$, and each slot $k \in [K]$ has an associated \emph{position bias} $\lambda_k \in (0,1]$. We let $\alpha_{k,t}=\alpha_{a_{k,t}}$ refer to the attractiveness parameter of item $a_{k,t}$. 

In response to the action $\mathbf{a}_t$, the user will either click on a displayed item or take a no click action. This process is captured via a click variable $Q_t$ taking values in $[K]_0$. The click probabilities follow from the MNL choice model, whose parameters are the products of attractiveness and bias parameters. Specifically, we have \begin{equation}
P(Q_t=k ~ | ~ \mathbf{a}_t) = \frac{\lambda_k \alpha_{k,t}}{1+ \sum_{v=1}^K \lambda_v\alpha_{v,t}}, \quad k \in [K], \label{eq::MNLclick}
\end{equation} and $P(Q_t=0 ~|~ \mathbf{a}_t)=1-\sum_{k=1}^K P(Q_t=k~|~ \mathbf{a}_t)$.

Following the user's choice, the decision-maker receives a reward $R(\mathbf{a}_t)= \mathbb{I}\{Q_t \neq 0\}$. The decision-maker's aim is to maximise their expected cumulative reward over $T$ rounds. The expected reward on an action $\mathbf{a} \in \mathcal{A}$ (in any round) is written, $r(\mathbf{a}) := \mathbb{E}(R(\mathbf{a})) = \sum_{k=1}^K P(Q=k ~ | ~ \mathbf{a})$.  The challenge for the decision maker is that the attractiveness parameters are unknown and the optimal action is therefore initially unclear. We will consider the problem in two informational settings: where the position biases are known, and unknown.

\subsection{On Approaches to the Problem}

The decision-maker faces a classic exploration-exploitation dilemma and must employ a strategy which balances between reward maximising and information gaining actions. We refer to such a strategy as a \emph{policy}, and formalise it as a (possibly randomised) mapping from a history $\mathcal{H}_{t-1}=\sigma(\mathbf{a}_1,Q_1,\dots,\mathbf{a}_{t-1},Q_{t-1})$ to an action $\mathbf{a}_t \in \mathcal{A}$ for each time $t \in [T]$.

We propose \emph{upper confidence bound} (UCB) policies for both the known and unknown position bias settings. UCB approaches are well-studied in the context of multi-armed bandits, following from \cite{LaiRobbins1985}, and are known to achieve optimal regret in a variety of settings. The canonical principle of a UCB approach is as follows. In each round, the policy computes high probability upper confidence bounds on the expected rewards of actions by utilising tight concentration results, and selects an action with maximal associated bound. Intuitively speaking, these approaches are effective because they tend to select actions that either have high reward (and thus are profitable) or high uncertainty (and thus provide a substantial information gain).

In the MNL-LTR setting, the identification of tight concentration results is the most involved aspect of designing UCB policies. In part, this is because the likelihood induced by the MNL model \eqref{eq::MNLclick} has a complex combinatorial structure, making it hard to identify parameter estimates with known distributional properties. Our proposed strategies subvert this issue by utilising a restriction on the decision-maker's actions such that the likelihood factorises usefully. This technique (first used by \cite{AgrawalEtAl2017,AgrawalEtAl2019}) restricts the decision-maker to repeatedly display any selected ordered list in each round until a no-click event is observed. Unbiased estimators may then be constructed as a sum of geometrically distributed random variables which are functions of the users' stochastic behaviour.

We will be interested in the empirical and theoretical performance of policies, measured in terms of their expected pseudoregret (referred to simply as regret in what follows) in $T$ rounds, defined as, \begin{equation}
Reg(T)= Tr(\mathbf{a}^*)- \mathbb{E}\bigg(\sum_{t=1}^T R(\mathbf{a}_t)\bigg), \label{eq::regret}
\end{equation} where $\mathbf{a}^* = \max_{\mathbf{a}\in\mathcal{A}}r(\mathbf{a})$ is an optimal action. Specifically, we will be interested in the order (w.r.t. $T,K,$ and $J$) of upper bounds on the regret for our proposed policies, and lower bounds on the regret which hold uniformly across all (reasonable) policies. We will study the problem in two informational settings: one where only attractiveness parameters are unknown, and another where both the attractiveness parameters and position biases are unknown. Our results establish an $\Omega(\sqrt{JT})$ lower bound on regret, and an upper bound matching this up to logarithmic factors for the known-position bias setting, and the first upper bound for the unknown-position bias setting of  $\tilde{O}(K^2\sqrt{JT})$. 

\subsection{Key Contributions}

The primary contributions of this work are threefold.  Firstly, we provide a new parametric model of LTR based on set-wise user decisions with foundations in classic choice theory. This model can capture the complex patterns of position effects in modern recommender systems, and the phenomenon where users consider multiple items simultaneously (not possible with factored click models).

Second, we derive new theoretical results concerning the concentration of Geometric random variables, giving rise to two new exponential inequalities: The first, an improved high-probability bound on the sum of non-independent, non-identically distributed (n.i.n.i.d.) Geometric random variables. The second, a high-probability bound on smooth functions of  n.i.n.i.d. Geometric random variables, which is applied to the maximum likelihood estimates (MLEs) in the unknown position bias setting to give non-asymptotic confidence sets for all parameters, even in the absence of closed-form expressions for the MLEs. 

Finally, based on these results we propose UCB algorithms for the known and unknown position bias settings, and validate their efficacy through derivation of upper and lower bounds on regret - which match up to logarithmic factors in the former setting - and empirical assessment against other state-of-the-art approaches.

\subsection{Related Work} \label{sec::lit}

A special case of our MNL-LTR model is the MNL bandit \citep{RusmevichEtAl2010}. It does not consider ordering of the items and coincides with our model when all position biases are equal. 

Initial studies of the MNL-bandit problem presented ``explore-then-commit" approaches, which only behave optimally for specific problem classes, and with prior knowledge of certain problem parameters \citep{RusmevichEtAl2010,SaureZeevi2013}.  \cite{AgrawalEtAl2019} and \cite{AgrawalEtAl2017} since presented UCB and TS approaches to the MNL-bandit respectively, which have $\tilde{O}(\sqrt{JT})$ regret, matching the minimax lower bound derived by \cite{ChenWang2018} (up to logarithmic factors). These methods use restrictions on decision-making to permit the construction of estimators with desirable properties. Such approaches and results are natural benchmarks\footnote{Independence from $J$ is possible in some variants - \cite{WangChenZhou2018} achieve this subject to further assumptions on the attractiveness parameters, and \cite{chen2021optimal,peeters2022stochastic} find such results for an `uncapacitated' version of the problem, but these are either less general or not-comparable to our model. Gap-dependent results have also been recently identified by \cite{yang2021fully}, however we retain a focus on minimax style results in this work.} for our problem, since when position bias $\lambda_k=1$ $\forall k\in[K]$ the MNL-LTR problem reduces to the MNL-bandit. 

When the position biases differ, however, these are not as successfully applied. As we show in the numerical experiments of Section \ref{sec::experiments} and the accompanying Appendix \ref{app::suboptimal}, an MNL-bandit algorithm can be applied to MNL-LTR by overparameterising the problem. That is to say, treating each item-slot combination as an `item' in its own right, placing constraints on which `items' can be selected concurrently, and mapping each selection to an ordered list. The theory of \cite{AgrawalEtAl2017,AgrawalEtAl2019} then guarantees $\tilde{O}(\sqrt{JKT})$ regret, but this is suboptimal, as it does not exploit the problem structure. Our approach to the known position bias setting does exploit this structure and thus enjoys a lower regret.

There has since been interest in extending the MNL-bandit model in various directions, considering the best-action identification variant \citep{ChenLiMao2018,fangfixed}, context-dependent variant \citep{OhIyengar2019, ChenWangZhou2018,bernstein2022exploration}, and variants models such as including variable rewards, inventory constraints, or omitting the `no-click' event \citep{BengsHullermeier2019, SahaGopalan2019, MesaoudiEtAl2020,dong2020multinomial,zhang2022mnl}. None of these works, however, fully consider the effect of position biases - i.e. as in our LTR variant - though they can be adapted to give consideration to ordering through specification of constraints.

Works on LTR are mainly distinguished by different click models \citep{ChuklinEtAl2015}, the majority being of the factored form previously described. Two notable choices are the Cascade Model (CM) \citep{CraswellEtAl2008} and the Position Based Model (PBM) \citep{RichardsonEtAl2007}. Under the CM the user considers each item in sequence, and decides whether or not to click on it before considering any items. If the user clicks an item, or reaches the end of the list without clicking any items, they stop. In contrast, under the PBM, the user may click on multiple items, and chooses whether to examine each item independently, with probabilities similar to our position biases. 

\cite{KvetonEtAl2015a} consider a LTR problem incorporating the CM, and \cite{LagreeEtAl2016} and \cite{KomiyamaEtAl2017} the PBM. In these specific settings upper confidence bound approaches can achieve $O(\sqrt{T})$ regret (optimal for those models), by exploiting the knoweledge of the true click model. 
Recently, \cite{gauthier2020position} have considered a version of PBM where the position effects are unknown - showing effective performance of a Thompson Sampling approach, but without theoretical guarantees. The works of \cite{ZoghiEtAl2017}, \cite{LattimoreEtAl2018}, and \cite{LiEtAl2019} have investigated more general click models which include the CM and PBM as special cases. The models of \cite{ZoghiEtAl2017} and \cite{LiEtAl2019} retain the assumption of a factored model, but are less restrictive than CM, and PBM. The model of \cite{LattimoreEtAl2018} makes sufficiently few assumptions to capture a wider range of models, including that which we propose. However such a general approach does not admit as tight theoretical guarantees. Table \ref{tab::comparison} compares the existing results on regret in LTR and the MNL-bandit with our regret upper bound.


\begin{table}[h]
\centering
\begin{tabular}{|l|c|c|c|c|}
\hline
 & MNL choice & LTR & Algorithm & Regret on MNL-LTR \\
\hline 
\cite{AgrawalEtAl2019} & $\checkmark$ & * & UCB & \multirow{2}{*}{$\tilde{O}\left(\sqrt{JKT}\right)$} \\
 \cite{AgrawalEtAl2017} & $\checkmark$ & * & TS & \\
 \cite{LattimoreEtAl2018} & included as special case & $\checkmark$ & TopRank & $O\left(\sqrt{JK^3 T}\right)$ \\
 \textbf{This paper} & $\checkmark$ & $\checkmark$ & UCB & $\tilde{O}\left(\sqrt{{JT}}\right)$ \\
\hline
\end{tabular}
\caption{Comparison of results in the present paper and related work for MNL-LTR with known position biases. $T$ denotes the number of rounds, $J$ the number of items, and $K$ the number of items chosen per round. An asterisk (*) in the LTR column denotes frameworks that can be adapted to LTR, but do so by over-parameterising the problem.}
\label{tab::comparison}
\end{table}

\section{Inference} \label{sec::knownPB}
In the MNL-LTR framework, the task of making accurate and efficient inference on the attractiveness parameters is more challenging than in other variants of LTR. Consider the likelihood of the sequence of clicks $Q_{1:T}=\{Q_1,\dots,Q_T\}$ given the attractiveness parameters $\bs\alpha$, position biases $\bs\lambda$, and action sequence $\mathbf{a}_{1:T}=\{\mathbf{a}_1,\dots,\mathbf{a}_T\}$, \begin{equation}
\mathcal{L}(Q_{1:T} ~|~ \mathbf{a}_{1:T},\bs\alpha,\bs\lambda) = \prod_{t=1}^T \frac{\sum_{j=0}^J \alpha_j \sum_{k=1}^K \lambda_k \mathbb{I}\{a_{k,t}=j,Q_t=k\}}{1+ \sum_{j=1}^J \alpha_j \sum_{k=1}^K \lambda_k \mathbb{I}\{a_{k,t}=j\}}. \label{eq::PBMlike}
\end{equation} The likelihood \eqref{eq::PBMlike} lacks a closed-form maximiser, meaning maximum likelihood estimators of $\bs\alpha$ and $\bs\lambda$ can only be computed numerically. Similarly, any Bayesian inference would necessarily be approximate, and computationally intensive. Both of these approximations (which are not necessary in related, factored models) are obstacles to the design and analysis of efficient, optimal sequential decision making policies.

\subsection{Inference with Known Position Biases} \label{sec::epochPB}

Exact inference is possible if we restrict the manner in which actions are selected. For the MNL-bandit, \cite{AgrawalEtAl2019} propose a restriction on decision-making that admits unbiased independent estimators of the attractiveness parameters. Specifically, if each selected set of items is displayed repeatedly until a no-click event is observed, then unbiased estimators of the attractiveness parameters are available. We will show that the same is possible in the MNL-LTR setting, if we display the same ranked list repeatedly until a no-click event occurs.

To describe this approach, we think of the $T$ rounds as being divided into $L \leq T$ epochs of variable length. An epoch $l \in [L]$ will consist of a sequence of consecutive time periods $\mathcal{E}_l \subseteq [T]$. In each epoch $l$ we will offer an ordered list $\mathbf{a}^l \in \mathcal{A}$ repeatedly, until a no-click event is observed. Let $a_k^l$ be the item in position $k$ in epoch $l$ and let $\alpha_k^l$ be the attractiveness parameter of this item, for $k \in [K]$.

As usual in each round $t \in \mathcal{E}_l$ a click variable $Q_t$ is observed. For each slot $k \in [K]_0$ the number of clicks on position $k$ in epoch $l$ is defined as $n^l_{k} = \sum_{t \in \mathcal{E}_l} \mathbb{I}\{Q_t=k\}$. By the construction of the epochs we always have $n_0^l=1$, unless $l=L$ and the final epoch is stopped by the completion of the time horizon, rather than a no-click event. We now show that these counts $n^l_k$ can be used to construct simple closed-form estimators of the attractiveness parameters.

The log-likelihood of the observed clicks $\mathbf{n}^l=(n_0^l,n_1^l,\dots,n_K^l)$ in a single epoch $l$, with fixed action $\mathbf{a}^l$ can be written as, \begin{displaymath}
\log\mathcal{L}(\mathbf{n}^l|\mathbf{a}^l,\bs\alpha) = \sum_{k=0}^K n_k^l \bigg[\log(\lambda_k\alpha_k^l)-\log(1+\sum_{v=1}^K \lambda_v\alpha_v^l) \bigg].
\end{displaymath} The single-epoch likelihood is maximised by estimators $\hat\alpha_k^l = {n_k^l}/{\lambda_k}$ for $k \in [K]$. 

Inspired by these within-epoch estimators, we may then construct estimators for each attractiveness parameter $\alpha_j$, $j \in [J]$, aggregating over $L$ complete epochs as \begin{equation}
\bar{\alpha}_j(L) = \frac{\sum_{l=1}^L \sum_{k=1}^K \mathbb{I}\{a_k^l=j\} n_k^l}{\sum_{l=1}^L \sum_{k=1}^K \mathbb{I}\{a_k^l=j\} \lambda_k}, \quad j \in [J]. \label{eq::unbiasedestimators}
\end{equation}  These estimators result from weighted averaging of the within-epoch maximum likelihood estimators - which is preferable to uniform averaging as we should expect epochs where the item $j$ was placed in a slot with a higher position bias to be less variable and thus more reliable. The lemma below, whose proof is reserved for Appendix \ref{app::technical_proofs}, gives the distribution of the random variables $n^l_k$. It follows immediately that our estimators $\hat\alpha_k^l$ and $\bar\alpha_j(L)$ are unbiased.

\begin{lemma} \label{lem::Geometric} For each $k \in [K]$, and $l \in [L]$, $n_k^l$, the number of clicks on the item in position $k$ during epoch $l$,  follows an Geometric\footnote{For clarity, we note that throughout this paper we use the following parametrisation of the geometric distribution. If $X \sim Geom(p)$ then $P(X=x)=(1-p)^x p$, $x \in \mathbb{N}:=\{0,1,\dots\}$.} distribution with parameter $(1+\lambda_k\alpha_k^l)^{-1}$.
\end{lemma}

\subsection{Inference with Unknown Position Biases} \label{sec::unknownPB}

When the position biases are unknown, epoch-based decision making is also useful. In this setting the likelihood is not identified unless we fix one of the position biases, so we fix $\lambda_1=1$. This restriction may rescale other parameters, with respect to the known position bias case, but crucially it does not change the interpretation of the model. Some further notation is also useful to describe inference in this setting. Define the $K \times J$ matrix of click counts in $l \in [L]$ epochs as $\mathbf{N}(l)$ having entries, \begin{displaymath}
N_{kj} = \sum_{l=1}^L \sum_{t \in \mathcal{E}_l}\mathbb{I}\{Q_t=k,\mathbf{a}_k^l=j\}, \quad k \in [K],~ j \in [J].
\end{displaymath} Similarly, define $\tilde{\mathbf{N}}(l)$ as the matrix of counts of selections of item-position combinations, whose entries are \begin{displaymath}
\tilde{N}_{kj} = \sum_{l=1}^L \mathbb{I}\{\mathbf{a}_k^l =j\}, \quad k \in [K],~ j \in [J].
\end{displaymath}  

Now, define $\gamma_{jk} = \alpha_j\lambda_k,$ $j \in [J],$ $k \in [K]$ to be the products of the attraction probabilities and position biases. Using the known distribution of the click counts $n_k^l$ we can derive an unbiased product parameter estimate $\bar\gamma_{jk}(L) = N_{kj}/\tilde{N}_{kj}$ for each $j \in [J],$ and $k \in [K]$. A naive approach would independently estimate the $JK$ product parameters and build UCBs around those. Such an approach does not make efficient use of the data, and as such associated decision-making rules can spend a prohibitively long time exploring, although they will eventually converge to optimal actions. We discuss the limitations of such an approach in more detail in Appendix \ref{app::suboptimal} and revisit it in the experiments in Section \ref{sec::experiments}. In the remainder of this section we will focus on direct inference on the attractiveness parameters and position biases.

We may obtain estimates of the attractiveness parameters and position biases via the EM scheme outlined in Algorithm \ref{alg::EMinf}. In particular this algorithm exploits that conditioned on estimates of the attractiveness parameters $\hat\alpha_{1:J}(L)$ we have an  estimate of the position bias for slot $k \in \{2,\dots,K\}$ as \begin{equation}
\hat\lambda_k(L) = \frac{1}{L}\sum_{j=1}^J \frac{\hat\gamma_{jk}(L)}{\hat\alpha_j(L)}\sum_{l=1}^L \mathbb{I}\{\mathbf{a}_k^l=j\} = \frac{1}{L}\sum_{j=1}^J \frac{N_{kj}}{\hat\alpha_j(L)}. \label{eq::EMlambda}
\end{equation}
Similarly, we have an estimate of the attractiveness of item $j \in [J]$ given estimates $\hat\lambda_{2:K}(L)$ of the position biases, as \begin{equation}
\hat\alpha_j(L) = \frac{1}{\sum_{l=1}^L\sum_{k=1}^K \mathbb{I}\{ \mathbf{a}_k^l=j\}}\sum_{k=1}^K \frac{N_{kj}}{\hat\lambda_k(L)}, \label{eq::EMalpha}
\end{equation} where $\hat\lambda_1(L)=\lambda_1=1$. Algorithm \ref{alg::EMinf} iterates between estimating position biases and attractiveness parameters until the estimates converge to within some tolerance.\footnote{There is an issue with the numerical stability of this EM scheme, as if a given item or position has no associated clicks, its estimate will go to 0. We can resolve this either by adopting the convention that $0/0=0$ or by artificially constraining the estimates to be no smaller than some $\epsilon>0$} The following lemma guarantees the convergence of this EM scheme. Its proof is given in Appendix \ref{app::technical_proofs}, and follows from the unimodality of the log-likelihood function. 

\begin{lemma} \label{lem::EMconv}
The estimators $\bs\alpha^{EM}$, and $\bs\lambda^{EM}$ derived from the EM algorithm, Algorithm \ref{alg::EMinf}, converge monotonically to the maximum likelihood estimators.
\end{lemma}

\begin{algorithm}[htbp]
    \caption{EM Algorithm for MNL-LTR with Unknown Position Biases}
    \label{alg::EMinf}
    \vspace{0.2cm}
    \textbf{Inputs:} Initial parameter values $\alpha_{j,0}$ for all $j \in [J]$, and $\lambda_{k,0}$ for $k \in \{2,\dots,K\}$. Tolerance parameter $0<\xi< 1$. Action and click histories, $\mathbf{a}^{1:L}$, $Q_{1:T}$.
    
    \vspace{0.2cm}
    
    Set $d \leftarrow 1$, $s \leftarrow 0$, and $\lambda_{1,t} \leftarrow 1$ for all $t \geq 0$.

     \vspace{0.2cm}
    
    \textbf{While} $d > \xi$ \textbf{do:}
    
    \begin{itemize}
    \item Set $s \leftarrow s +1$.
\item \textbf{E-Step} For each $k \in \{2,\dots,K\}$, calculate $\lambda_{k,s}$ according to \eqref{eq::EMlambda}.
\item \textbf{M-Step} For each $j \in [J]$, calculate $\alpha_{j,s}$ according to \eqref{eq::EMalpha}.
\item Calculate $d=\max \left(\max_{k \in \{2,\dots,K\}} | \lambda_{k,s}- \lambda_{k,s-1}|, \max_{j \in [J]} |\alpha_{j,s}-\alpha_{j,s-1}|\right)$.
\end{itemize}
	\textbf{Return} $\lambda_{1,s},\dots,\lambda_{K,s}$ and $\alpha_{1,s}\,\dots,\alpha_{J,s}$ as estimates of position biases and attractiveness parameters.

        \vspace{0.2cm}
\end{algorithm}

\section{Concentration Results} \label{sec::conc_theory}

In this section we derive concentration results for the parameter estimates in both the known and unknown position bias settings. Quantification of the uncertainty in the parameters is key to designing effective sequential decision-making algorithms, and the results in this section will later be used to construct UCB approaches.

\subsection{Concentration Results Relevant to the Known Position Bias Setting} \label{sec::concUCB}
As discussed in Section \ref{sec::epochPB}, the empirical means, $\bar\alpha_{j}$, are weighted averages of Geometric random variables. The following theorem gives a martingale-type concentration result for the sum of geometrically distributed random variables with differing means. This result is not specific to the MNL-LTR or MNL bandit settings, and therefore may be of independent interest. 

It is worth noting that the results of Theorem \ref{lem::MartConc} simultaneously have improved coefficients, and a greater generality than alternative results for i.i.d. geometric random variables obtained by \cite{AgrawalEtAl2019}. We require the greater generality in the MNL-LTR setting because the random variables associated with clicks of an item per epoch will be a) non-identically distributed as they depend on the position bias, and b) non-independent as the assignment of items to slots depends on the previously observed data.

\begin{theorem} \label{lem::MartConc}
Consider geometric random variables $Y_i$ with parameter $p_i$, $i \in [n]$, where $p_i$ may be a function of $p_1,\dots, p_{i-1},Y_1, \dots Y_{i-1}$. Let $\mu_i=\frac{1-p_i}{p_i}$, and $\sigma_i^2 = \mu_i^2+\mu_i$. If $\mu_i \leq 1$ for all $i\in [n]$, then we have for all  $C>0$, \begin{align}
P \left( \left|{\sum_{i=1}^n Y_i} - {\sum_{i=1}^n \mu_i} \right| > \sqrt{2\sum_{i=1}^n\sigma_i^2\log(C)} + {4\log(C)}\right) &\leq 2C^{-1}, ~\forall n \geq 1. \label{eq::MartConcpart1}
\end{align} Furthermore, we have for all $C>0$, 
\begin{align}
P \left( \left|{\sum_{i=1}^n Y_i} - {\sum_{i=1}^n \mu_i} \right| > \sqrt{8\sum_{i=1}^n Y_i\log(C)} + {4\log(C)}~ \bigg| ~A_n\right) &\leq 4C^{-1}, \label{eq::MartConcpart2}
\end{align} where $A_n = \left\lbrace \sum_{i=1}^n \mu_i \geq 8\log(C) + \sqrt{8\sum_{i=1}^n \sigma_i^2\log(C)} \right\rbrace$.
\end{theorem}
A full proof of Theorem \ref{lem::MartConc} is provided in Appendix \ref{app::martingale}, but we briefly outline its intuition here. The proof derives a new bound on the central moments of the geometric distribution in order to utilise a Bernstein-like inequality for martingale difference sequences. As the central moments of the geometric distribution lack a closed-form expression, this is non-trivial. We achieve the bound by first bounding the cumulants of the geometric distribution and exploiting a combinatorial link between central moments and cumulants.

The following lemma adapts the result of Theorem \ref{lem::MartConc} to the LTR setting. Its proof is also given in Appendix \ref{app::martingale}. The UCB algorithm we propose in Section \ref{sec::algorithms} for the known position bias setting is designed to exploit these results.

\begin{lemma} \label{lemma::epochUCBconc}
We have for estimators $\bar\alpha_{j}(l)$ $j \in [J]$ defined as in Equation \eqref{eq::unbiasedestimators}, and attractiveness parameters $0<\alpha_j \leq 1$, $j \in [J]$, the following concentration results, for all $l: \Lambda_{j,l}>0$
\begin{align}
P \left( \left|\bar\alpha_{j}(l) - \alpha_j\right| > \sqrt{\frac{4\alpha_j\log(Jl^2/2)}{\Lambda_{j,l}}} + \frac{{4\log(Jl^2/2)}}{\Lambda_{j,l}}\right) &\leq \frac{4}{Jl}, \label{eq::UCBConcpart1}\\
P \left( \left|\bar\alpha_{j}(l) - \alpha_j\right| > \sqrt{\frac{8\bar\alpha_{j}(l)\log(Jl^2/2)}{\Lambda_{j,l}}} + \frac{{8\log(Jl^2/2)}}{\Lambda_{j,l}}\right) &\leq \frac{6}{Jl}. \label{eq::UCBConcpart2}
\end{align} Furthermore, for $l: \Lambda_{j,l}> 4\log(Jl^2/2)/\alpha_j$ we have,
\begin{align}
P \left(\bar\alpha_j(l) > 2\alpha_j + \frac{4\log(Jl^2/2)}{\Lambda_{j,l}}\right) &\leq \frac{2}{Jl}. \label{eq::UCBconcpart3}
\end{align}
\end{lemma}

\subsection{Concentration Results Relevant to the Unknown Position Bias Setting} \label{sec::concUCBunknown}

The derivation of concentration results in the unknown position biases setting is more challenging, since the MLE for any unknown parameter (attractiveness or position bias) does not have a closed form. The asymptotic properties of MLEs are well documented, but there are comparatively few general guarantees relating to finite-time behaviour. However, here we are able to utilise non-asymptotic deviation inequalities on certain functions of random variables to derive concentration properties for a family of MLEs derived from geometrically distributed data. 

As in the previous section we have a general concentration result, followed by an application to the MNL-LTR setting. 
We begin with Theorem \ref{thm::MVgeneral}, whose proof is given in Appendix \ref{app::MLconc} which gives a  deviation inequality for a function of multivariate Geometric data. The derivation of this result is based on theory from \cite{BobkovLedoux1998} and a logarithmic Sobolev inequality of \cite{JoulinPrivault2004}. Before stating our result we introduce a notion of the smoothness of a discrete function expressed in terms of its finite differences.

Let $(\epsilon_{li})_{i \in [d],l \in [n]}$ denote the canonical basis on $\mathbb{R}^{d \times n}$. For a function $f: \mathbb{N}^{d \times n} \rightarrow \mathbb{R}$ define the finite difference with respect to the input variable indexed $l,i$,  \begin{align*}
D_{li}f(\mathbf{X}) = F(\mathbf{X}+\epsilon_{li})-F(\mathbf{X}), \quad \mathbf{X} \in \mathbb{N}^{d \times n}.
\end{align*}  We say that a function $F: \mathbb{N}^{d\times n}\rightarrow \mathbb{R}$ is $(\beta_1,\beta_2)$-smooth, for parameters $\beta_1,\beta_2>0$, if, \begin{align}
\sum_{l=1}^n \sum_{i=1}^{d}|D_{li}F|^2 \leq \beta_1^2, \quad \text{and} \quad 
\max_{l \in [n]} \max_{i \in [d]} \left(|D_{li}F|\right)\leq \beta_2 \quad \forall ~ \mathbf{X} \in \mathbb{N}^{d \times n}. \label{eq::quadraticFDGbound} 
\end{align}

\begin{theorem} \label{thm::MVgeneral}
Let $n,d \in \mathbb{N}$ and $\mu_l ~|~ \mu_{1:l-1}$ $l \in [n]$ be a series of conditional multivariate geometric measures on $\mathbb{N}^{d}$, such that each component, $\mu_{li}$ is a geometric law with parameter $p_{li} \in (0,1]$ for $i \in [d]$, $l \in [n]$. Define $\mu^n = \bigotimes_{l=1}^n \mu_l$ as the product measure, and let $F$ be a $(\beta_1,\beta_2)$-smooth function with $\beta_1>0$, and $\beta_2 \in (0, \max_{i,l} ( - \log(1-p_{li}))]$. Then $\mathbb{E}_{\mu^n}(|F|)<\infty$, and for every $\delta>0$, \begin{equation}
\mathbb{P}_{\mu^n}\left(F \geq \mathbb{E}_{\mu^n}(F) + \delta\right) \leq \exp\left(\min \left\lbrace \frac{-\delta^2}{4\beta_1^2M}, \frac{(\log(1-p))^2\beta_1^2M}{4\beta_2^2} + \frac{\log(1-p)\delta}{2\beta_2}\right\rbrace \right),
\end{equation} where $M>0$ is a known finite constant depending on the parameters $\{p_{li}\}_{i \in [d],l\in [n]}$.
\end{theorem} 

Choosing the function $F$ to be the MLEs in the MNL-LTR setting, we have the following result, giving concentration inequalities for the estimators derived from Algorithm \ref{alg::EMinf}.

\begin{corollary} \label{corr:MLEconc}
We have for EM estimators $\alpha_{l,j}^{EM}$, $j \in [J]$ and attractiveness parameters $0< \alpha_j \leq 1$, $j \in [J]$, that for all $l: \sum_{k=1}^K \tilde{N}_{kj}>0$, \begin{align*}
P\left(\left|\alpha^{EM}_{l,j}- \alpha_j \right| > \sqrt{36\beta_{1,l,j}^2\log(Jl^2)} \right) &\leq \frac{2}{Jl^2},
\end{align*} 
where $\beta_{1,j,l}$ is a sensitivity parameter defined as \begin{align*}
    \beta_{1,l,j} &:= \sqrt{\sum_{k,s: \tilde{N}_{ks}\geq 1} \left(\alpha_{l-1,j}^{EM} - \alpha_{l-1,ks,j}^{EM}\right)^2\tilde{N}_{ks}}, \text{ for } j \in [J], l\geq 1.
\end{align*}
\end{corollary}
The proof of this corollary is reserved for Appendix \ref{app::MLconc}. Its main argument is to recognise that the restriction of $\alpha_l^{EM}$ to its $j^{th}$ output, for fixed selections matrix $\tilde{\mathbf{N}}(l)$, is (subject to a minor rearrangement of the inputs) a function from $\mathbb{N}^{K \times l} \rightarrow [0,1]$, to which the functional inequality of Theorem \ref{thm::MVgeneral} applies.

A similar result holds for the position bias parameters but is not required. This is since the algorithm we propose in the following section does not need to construct UCBs for the position biases as every slot is utilised in every round.

Finally, for the purpose of the regret bounds which will follow in Section \ref{sec::theory}, Lemma \ref{lem::betabound}, below, relates the sensitivity parameter $\beta_{1,l,j}$ to the number of selections of item $j$ in $l$ epochs. Its proof is given in Appendix \ref{app::MLconc}, and is based on bounding inverse gradients of the log-likelihood to quantify sensitivity of the MLEs.

\begin{lemma} \label{lem::betabound}
    There exists a constant $C>0$ independent of $J, K,$ and $L$ such that the sensitivity parameter $\beta_{1,j,l}$ satisfies $$\beta_{i,l,j} \leq C\sqrt{\frac{JK}{\max_k \tilde{N}_{kj}}}.$$
\end{lemma}

\section{Decision-making Algorithms} \label{sec::algorithms}

We now outline our new UCB approaches. As is typical, the algorithms select actions which maximise the expected reward with respect to a set of upper confidence bounds on the attractiveness parameters. Such an optimal action will place the item with the $k^{th}$ largest UCB in the slot with the $k^{th}$ largest position bias (or estimated position bias if position biases are unknown) for each $k \in [K]$. Then, however, the algorithms will repeatedly use this action in each round until a no-click event is observed. This is in contrast to the traditional approach of calculating new UCBs in every round.

\begin{algorithm}[]
    \caption{Epoch-UCB algorithm for known position biases}
    \label{alg::PBEpochUCB}
    \vspace{0.2cm}    
     Initialise with $l=0$, and $Q_0=0$. Iteratively perform the following for $t \in [T]$,  \vspace{0.2cm}
     
     If $Q_{t-1}=0$ \begin{itemize}
     \item Set $l \leftarrow l+1$
     \item Calculate UCBs. For $j \in [J]$ compute,  $$\alpha_{j,l}^{UCB}= \bar\alpha_{j}(l-1)+ \sqrt{\frac{4\min(1,2\bar\alpha_{j}(l-1))\log(Jl^2/2)}{\Lambda_{j,l-1}}} + \frac{{4\log(Jl^2/2)}}{\Lambda_{j,l-1}}.$$
     \item Select an action $\mathbf{a}_t \in \argmax_{\mathbf{a} \in \mathcal{A}} r_{\bs\alpha_{l}^{UCB}}(\mathbf{a})$ which is optimal with respect to the UCB vector $\bs\alpha_l^{UCB}:=(\alpha_{1,l}^{UCB},\dots,\alpha_{l,J}^{UCB})$, and observe click variable $Q_t$
     \end{itemize}
     otherwise, set action $\mathbf{a}_t=\mathbf{a}_{t-1}$, and observe click variable $Q_t$.
        \vspace{0.2cm}
\end{algorithm}

In Algorithm \ref{alg::PBEpochUCB} we present our Epoch-UCB method for the variant of MNL-LTR with known position biases. In each epoch $l \in [L]$ the algorithm computes a UCB, $\alpha_{j,l}^{UCB}$ for each item $j \in [J]$. This UCB is constructed using the concentration results of Lemma \ref{lemma::epochUCBconc} to give an upper bound on $\alpha_j$ with high probability. The $\min(1,2\bar\alpha_{j,l-1})$ term allows the UCB to adapt to whichever of \eqref{eq::UCBConcpart1} and \eqref{eq::UCBConcpart2} gives the tighter bound. Henceforth, for $l\geq 1,$ and $ j\in [J]$, we define $\Lambda_{j,l} := \sum_{k=1}^K \sum_{s=1}^l \lambda_k \mathbb{I}\{\mathbf{a}_k^s =j\}$.

To state our algorithm for the setting where position biases are unknown define $\alpha^{EM}: \mathbb{N}^{K\times J} \times \mathbb{N}^{K \times J} \rightarrow [0,1]^J$ be the function which takes click and selection count matrices as inputs and returns the EM estimates for attractiveness parameters $\bs\alpha$. In effect, $\alpha^{EM}$ represents the application of Algorithm \ref{alg::EMinf}.

\begin{algorithm}[]
    \caption{Epoch-UCB algorithm for unknown position biases}
    \label{alg::UPBEpochUCB}
    \vspace{0.2cm}    
    
    Initialise with $l=0$ and $Q_0=0$. Iteratively perform the following for $t \in [T]$, \vspace{0.2cm}
    
     If $Q_{t-1}=0$ \begin{itemize}
     \item Set $l \leftarrow l+1$. Compute EM estimators and finite difference bounds, \begin{align}
     \bs\alpha_{l}^{EM} &= \alpha^{EM}(\mathbf{N}(l-1),\tilde{\mathbf{N}}(l-1)) \label{eq::standardMLE} \\
     \bs\alpha_{l,kj}^{EM} &= \alpha^{EM}(\mathbf{N}(l-1) + \bs\epsilon_{kj},\tilde{\mathbf{N}}(l-1)), \enspace \forall k,j: \tilde{N}_{kj}\geq 1 \label{eq::modifiedMLE} \\
     \beta_{1,l,j}^2 &= \sum_{k,s: \tilde{N}_{ks}\geq 1} \left(\alpha_{l-1,j}^{EM} - \alpha_{l-1,ks,j}^{EM}\right)^2\tilde{N}_{ks}, \enspace \forall j \in [J] \label{eq::beta1} 
     \end{align}
     
     \item Calculate UCBs. For $j \in [J]$ compute,  $$\alpha_{j,l}^{UCB-U}= \alpha_{j,l}^{EM} + \sqrt{36\beta_{1,j,l}^2\log(Jl^2)}. $$
     \item Select an action $\mathbf{a}_t \in \argmax_{\mathbf{a} \in \mathcal{A}} r_{\bs\alpha_{l}^{UCB-U}}(\mathbf{a})$, which is optimal with respect to the UCB vector, and observe click variable $Q_t$.
     \end{itemize}
     otherwise, set action $\mathbf{a}_t=\mathbf{a}_{t-1}$, and observe click variable $Q_t$.
        \vspace{0.2cm}
\end{algorithm}

In Algorithm \ref{alg::UPBEpochUCB} we give our policy for the setting where position biases are not known. Its structure is similar to Algorithm \ref{alg::PBEpochUCB}, but the computation of the UCBs is more involved as it involves finite difference gradients. In each epoch $l \in [L]$, estimates of the MLEs, $\bs\alpha_l^{EM}$, are computed via Algorithm \ref{alg::EMinf} as in \eqref{eq::standardMLE}. Our approach proceeds to calculate further estimates of the attractiveness parameters but on modified data, as in \eqref{eq::modifiedMLE}. For each item-position pair that has been selected at once, i.e. each $k,j$ with $\tilde{N}_{kj} \geq 1$, we compute $\bs\alpha_{l,kj}^{EM}$, parameter estimates based on $l$ epochs of data but with $N_{kj}$ incremented by 1. A sum of the squared finite differences is then computed as in \eqref{eq::beta1}, which is used in the UCB inspired by Corollary \ref{corr:MLEconc}. This is in place of a supremum bound on the sum of squared differences over all possible outcomes, which would be difficult to compute in practice.

\section{Regret Bounds} \label{sec::theory}

In this section we give upper and lower bounds on the regret for MNL-LTR algorithms. Proposition \ref{thm::PBregret} gives our upper bound on the regret incurred by Algorithm \ref{alg::PBEpochUCB} when the position biases are known. Proposition \ref{thm::lb} gives a lower bound on the regret of any algorithm, in terms of $S_K=\sum_{k=1}^K \lambda_k$, and $S_{K,2}=\sum_{k=1}^K \lambda_k^2$. The proofs of both results are given in the appendix - Proposition \ref{thm::PBregret} in Appendix \ref{app::regret}, and Proposition \ref{thm::lb} in Appendix \ref{app::lb}.

\begin{proposition} \label{thm::PBregret}
The regret in $T$ rounds of the the Epoch-UCB approach, Algorithm \ref{alg::PBEpochUCB}, for any MNL-LTR problem where the item attractiveness parameters satisfy $\alpha_j \leq \alpha_0 =1$, $j \in [J]$ and the position biases $\lambda_{k} \leq 1$, $k \in [K]$ are known satisfies \begin{displaymath}
Reg(T) = O \left(\sqrt{\frac{\max_{k\in[K]}\lambda_k\log(JT)JT}{\min_{k \in [K]}\lambda_k}} \right).
\end{displaymath} 
\end{proposition}

\begin{proposition} \label{thm::lb}
The regret of any algorithm for the MNL-LTR problem with position biases $1 \geq \lambda_1 > \lambda_2 > \dots > \lambda_K>0$ satisfying $S_{K,2}>1$ and $J\geq 4K$ items with attractiveness parameters $\alpha_j \in (0,1]$, $j\in [J]$, is lower bounded as \begin{equation}
Reg(T) = \Omega\left(\sqrt{\frac{JTS_{K,2}}{S_{K}}}\right).
\end{equation}
\end{proposition} 

The upper and lower bounds in the known position bias case match in their order with respect to $J$ and $T$ (up to logarithmic factors). 

For the case of unknown position biases, the complex form of the derived UCB algorithm prohibits an analysis which as sharply quantifies the dependence on position biases. Nevertheless, we have an upper bound on the regret of Algorithm \ref{alg::UPBEpochUCB} in terms of $J,K,$ and $T$, in the proposition below, whose proof is in Appendix \ref{app::regret}.

\begin{proposition} The regret in $T$ rounds of the Epoch-UCB approach, Algorithm \ref{alg::UPBEpochUCB}, for any MNL-LTR problem where the unknown item attractiveness parameters satisfy $\alpha_j \leq \alpha_0 =1$, $j \in [J]$ and unknown position biases satisfy $\lambda_{k} \leq 1$, $k \in [K]$ satisfies \begin{displaymath}
Reg(T) = O \left(K^2\sqrt{{\log(JT^2)JT}} \right).
\end{displaymath} \label{prop::UPBregret}
\end{proposition}
There is some suboptimality in the bound of Proposition \ref{prop::UPBregret} with respect to $K$. We believe this arises, in part, as an artefact of the (simplifying) focus on a single $k=\argmax \tilde{N}_{kj}$ in Lemma \ref{lem::betabound}, and it may yet be possible (through further laborious analytic work) to achieve a less elegant but tighter bound.

\section{Experiments} \label{sec::experiments}

We now conduct empirical comparisons on three instances of MNL-LTR: \begin{enumerate}
\item[(a)] There are $K=4$ slots with position biases $\bs\lambda=(1,0.3,0.2,0.1)$. There are $J=6$ items with attractiveness parameters $\bs\alpha=(0.3,0.28,0.26,0.24,0.22,0.2)$.
\item[(b)] There are $K=3$ slots with position biases $\bs\lambda=(1,0.2,0.9)$. There are $J=4$ items with attractiveness parameters $\bs\alpha=(0.05,0.1,0.15,0.2)$. 
\item[(c)] There are $K=6$ slots with position biases $\bs\lambda=(1,0.9,0.7,0.3,0.5,0.7)$ and $J=30$ items, four having attractiveness parameter 1, two having attractiveness parameter 0.8, and the remaining twenty-four having attractiveness parameter 0.1.
\end{enumerate} Together, these cover several scales of problem where users may simultaneously consider multiple options with different prominence and items may be modelled as having independent attractiveness.

We consider both Algorithm \ref{alg::PBEpochUCB} which knows the position biases and Algorithm \ref{alg::UPBEpochUCB} where the position biases are inferred. We will refer to the former as Epoch-UCB, and the latter as Epoch-UCB UPB (Unknown position biases) in what follows. Experimental results suggest that while the Epoch-UCB UPB algorithm does eventually learn the optimal actions, it can be overly conservative. We therefore also investigate a modification, Epoch-UCB* UPB, which is identical to Algorithm \ref{alg::UPBEpochUCB} except the UCB for item $j \in [J]$ in epoch $l \in [L]$ is calculated as $\alpha_{j,l}^{UCB} = \alpha_{j,l}^{EM} + 0.5\sqrt{\beta_{1,j,L}^2\log(\sqrt{J}l)}$.

We compare our algorithms to a range of alternative approaches. Firstly, we have a further known-position-bias epoch-based approach, Epoch-UCB-W. This uses the coefficients we would expect from adapting the weaker concentration inequalities used in \cite{AgrawalEtAl2019}. Epoch-UCB-W is identical to Algorithm \ref{alg::PBEpochUCB} except it calculates UCB index for item $j \in [J]$ in epoch $l \in [L]$ as: \begin{displaymath}
\alpha_{j,l}^{UCB-W} = \bar\alpha_{j,l-1} + \sqrt{\frac{48\min(1,2\bar\alpha_{j,l-1})\log(\sqrt{J}l/\sqrt{2})}{\Lambda_{j,l-1}}}+\frac{48\log(\sqrt{J}l/\sqrt{2})}{\Lambda_{j,l-1}}.
\end{displaymath} 

Second, we also consider the TopRank algorithm of \cite{LattimoreEtAl2018}. This algorithm can operate without knowledge of the position biases, but assumes that the slots are of decreasing attractiveness. TopRank has a markedly different structure to Epoch-UCB. TopRank maintains a hierarchical partition of the item set, such that the items sit in strata based on their perceived attractiveness. In each round the displayed list is constructed by randomising the order of the $n_1 \geq 1$ items in the top strata and assigning these to the first $n_1$ slots, then randomising the order of the $n_2 \geq 1$ items in the second strata, assigning these to the next $n_2$ slots, and proceeding in such a fashion until all $K$ slots are filled. Items are demoted to lower strata if they have received sufficiently fewer clicks than another item in their strata. 

As discussed in Section \ref{sec::lit}, the other LTR approaches that we are aware of are all designed with factored click models in mind, and do not carry performance guarantees to the MNL-LTR setting. We do however investigate the Position Bias Upper Confidence Bound (PBUCB) algorithm of \cite{LagreeEtAl2016}, which is based on the position bias click model. This algorithm can use our position bias parameters in terms of its model, but will underestimate the $\alpha$ parameters as it expects that multiple items may be clicked by a user - i.e. its inference model is inconsistent with the MNL data generating process. Further modifications would be necessary to deploy a version of this algorithm if the position biases were not known.

Finally, we compare to the UCB approach described in Appendix \ref{app::suboptimal}. This approach, which we refer to as `MNL-bandit' in the figures, ignores some of the LTR structure, and treats the unknown position bias version of the problem as a constrained MNL bandit problem. It learns the product parameters $\gamma_{j,k}=\lambda_k\alpha_j$ individually and avoids the need for running the EM algorithm. As discussed in Appendix \ref{app::suboptimal} it does have sublinear regret guarantee, but must perform more exploration than other approaches due to being overparameterised.

\begin{figure}
\centering
\includegraphics[width=\textwidth]{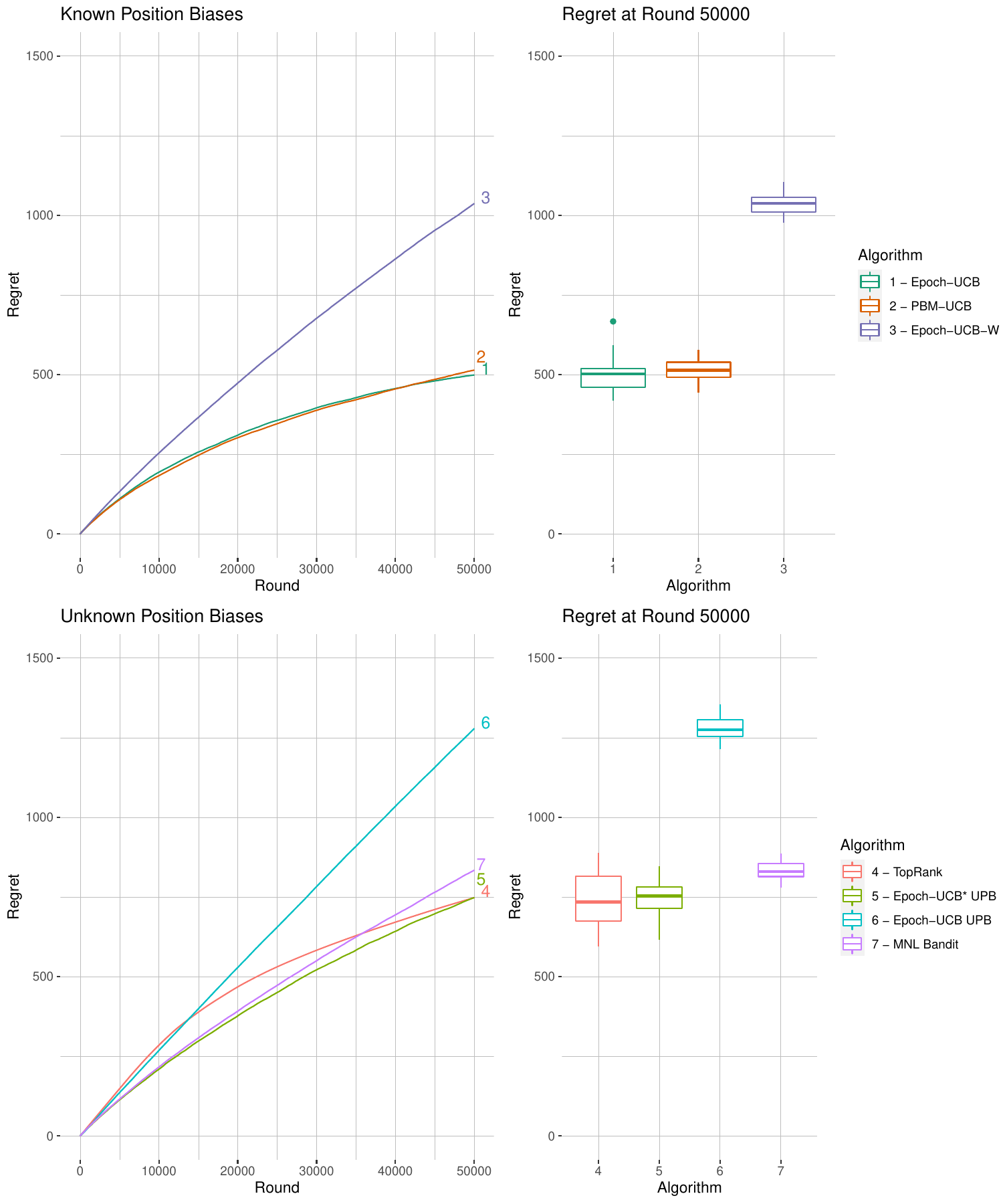}
\caption{Performance of algorithms on problem (a). The left panel shows the mean cumulative regret trajectory for each algorithm over 50000 rounds. The right panel shows the distribution of regret by algorithm at the end of 50000 rounds.}
\label{fig::regretA}
\end{figure}

Note that, problems (b) and (c) give examples where the optimal ordering of items is \emph{not} in decreasing order of attractiveness. In (b) for instance, the final slot, not the second slot, has the second-to-largest position bias. In the unknown position bias variant of this problem, Epoch-UCB UPB and Epoch-UCB* UPB can adapt to this as they actively learn the position biases, but algorithms assuming decreasing position bias, such as TopRank, cannot. 

\begin{figure}
\centering
\includegraphics[width=\textwidth]{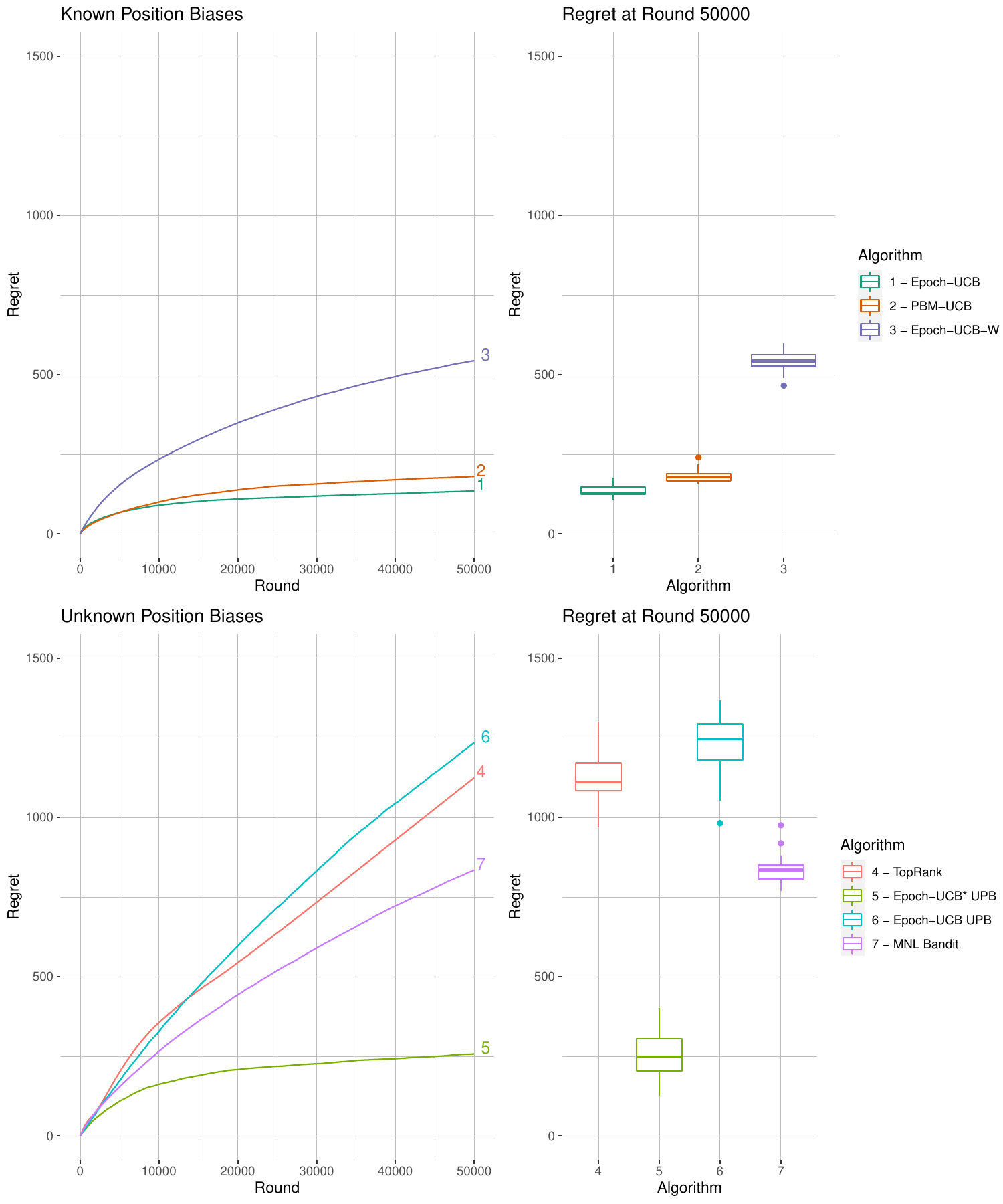}
\caption{Performance of algorithms on problem (b). The left panel shows the mean cumulative regret trajectory for each algorithm over 50000 rounds. The right panel shows the distribution of regret by algorithm at the end of 50000 rounds.}
\label{fig::regretB}
\end{figure}

\begin{figure}
\centering
\includegraphics[width=\textwidth]{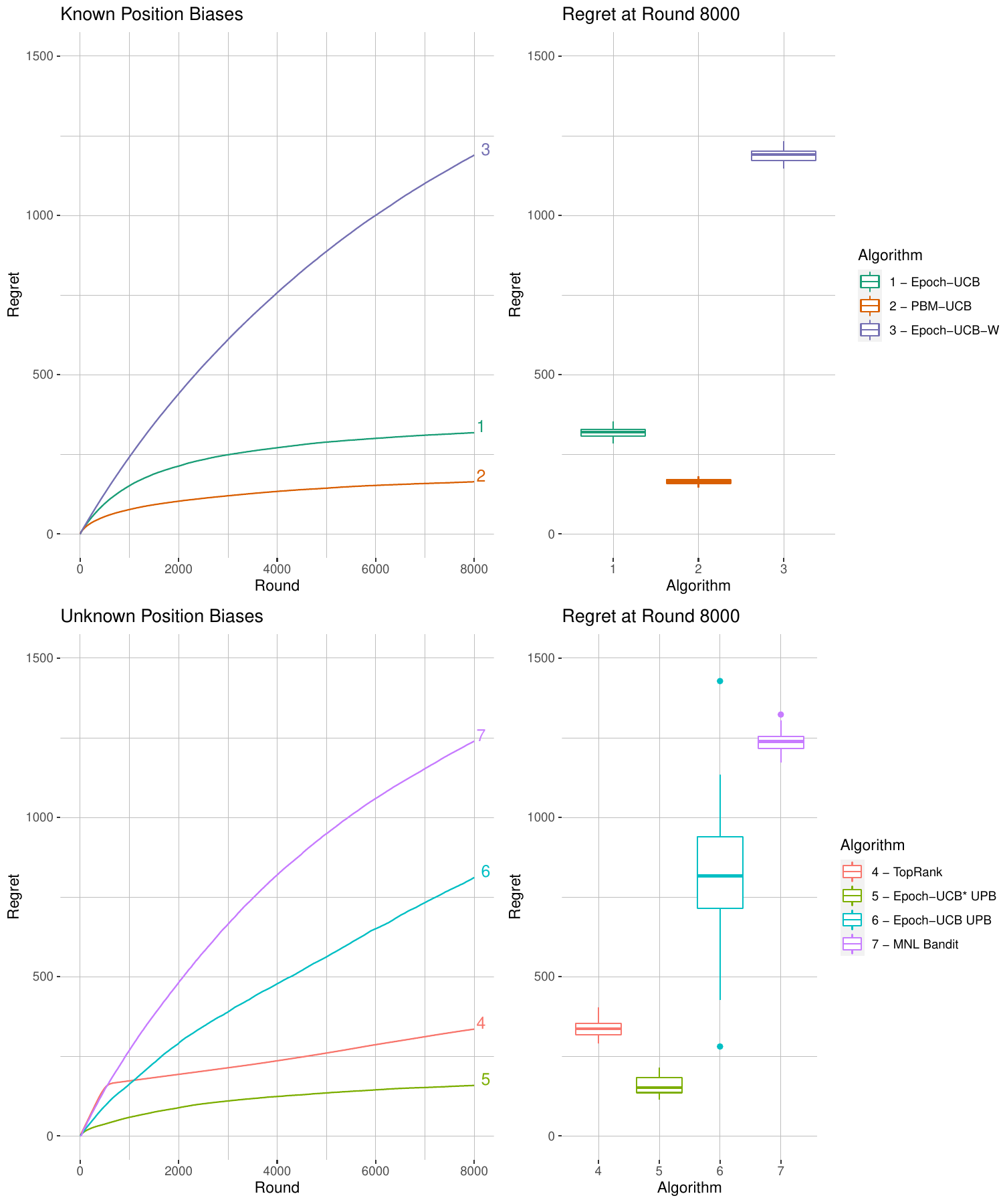}
\caption{Performance of algorithms on problem (c). The left panel shows the mean cumulative regret trajectory for each algorithm over 50000 rounds. The right panel shows the distribution of regret by algorithm at the end of 8000 rounds.}
\label{fig::regretC}
\end{figure}

The aforementioned algorithms were applied to problems (a) and (b) over 50000 decision-making rounds, over 40 replications. For problem (c) we use 8000 decision-making rounds, and 40 replications, since the optimal action can be learned more quickly. Figures \ref{fig::regretA}, \ref{fig::regretB}, and \ref{fig::regretC} display the results, in two forms: the mean regret accumulated through time in their left panes and the distribution of regret in the final rounds in their right. We focus only on the distribution in the final rounds as the results are such that plotting error bars with the each of the seven mean trajectories would make the graphs difficult to read.

Across all three problems we find that our Epoch-UCB and the PBUCB algorithms generally perform best. This is to be expected as they have access to the known position biases and assume a position based model. Our improved coefficients for the known position-bias setting are seen to have a substantial benefit as Epoch-UCB-W is much more conservative than Epoch-UCB. Indeed Epoch-UCB-W even suffers worse performance than TopRank which can identify the correct ordering of items despite not knowing the true click model. 

In the unknown position bias setting, we see that the unmodified Epoch-UCB UPB approach is also overly conservative, but the approach with modified coefficients Epoch-UCB* UPB is not much worse than the best known position bias algorithms. TopRank does not know the click model but performs well when the position biases are decreasing in the slot index. In problems (b) and (c) where the position biases do not fit this assumption TopRank can identify the top $K$ items reliably, but incurs a linear regret due to repeatedly ordering these suboptimally.

Problem (c) most clearly demonstrates the issue with the MNL-bandit approach. Here, $K$ and $J$ are much larger than in problems (a) and (b), and the MNL-bandit approach continues to explore long after Epoch-UCB* UPB has reached the point of mostly exploiting near-optimal actions. This is due to the fact that the MNL-bandit approach aims to collect data to estimate $JK=180$ different parameters, whereas Epoch-UCB* UPB utilises the known click model such it only estimates $J+K-1=35$ (assuming $\lambda_1=1$). This example displays that although the MNL-bandit approach has a sublinear regret guarantee, as shown in Appendix \ref{app::suboptimal}, it may be inappropriate in practice.

\section{Conclusion} \label{sec::discussion}

In this paper, we have proposed and analysed the multinomial logit choice variant of the learning to rank problem. Distinct from other model-based treatments of learning to rank, our model captures the behaviour of a user who makes a decision over a set of alternatives, rather than making a sequence of independent decisions. 

We proposed upper confidence bound approaches for the problem in two informational settings - where the effects of rank are known and unknown respectively. Both of these approaches are derived from new concentration theory. In the known position bias setting, we have derived a verified a Bernstein moment condition on the moments of the Geometric distribution and provided new martingale inequalities for geometric random variables. In the unknown position bias setting we have provided new functional inequalities for geometric data, giving concentration results for numerical maximum likelihood estimators. Further we have provided upper and lower bounds on regret in the known position bias setting, and simulations to display the effectiveness of our approaches.

Our proposed framework makes few assumptions beyond the MNL choice, which is a long-standing popular model in decision theory and may be applicable in numerous domains. Our concentration results are also of potential interest in other areas. Further, we lay a groundwork for further study of MNL-type LTR problems. Future work may consider randomised approaches, utilising similar posterior approximations as in \cite{AgrawalEtAl2017}, or recently proposed bootstrapping techniques as in \cite{KvetonEtAl2019}. The development of algorithms for a contextual variant of the problem, or with more complex or alternatively structured actions (perhaps bespoke to particular settings) would also seem to be worthy avenues to follow in the extension of this work.

\acks{The authors gratefully acknowledge the support of EPSRC grant EP/L015692/1 (STOR-i Centre for Doctoral Training).}

\bibliography{PostdocRefs.bib}

\appendix

\section{Proof of Geometric Martingale Concentration} \label{app::martingale}

In this section, we provide proofs of Theorem \ref{lem::MartConc} and Lemma \ref{lemma::epochUCBconc}, which comprise the concentration results relevant to the known position bias setting. The following result is a key component of the proof of Theorem \ref{lem::MartConc}. It gives a Bernstein-like bound for heavy-tailed martingale data.

\begin{lemma}[Theorem 1.2A of \cite{DeLaPena1999}] \label{lem::delaP}
Let $\{d_j,\mathcal{F}_j\}$ be a martingale difference sequence with $\mathbb{E}(d_j ~|~ \mathcal{F}_{j-1})=0$, $\mathbb{E}(d_j^2 ~|~ \mathcal{F}_{j-1})=\sigma^2_j$, for each $j$, and $V_n^2=\sum_{j=1}^n \sigma_j^2$. Furthermore assume that \begin{equation}
\mathbb{E}(|d_j|^k ~|~ \mathcal{F}_{j-1}) \leq \frac{k!}{2} \sigma_j^2 c^{k-2} \enspace a.e \label{eq::delapenacondition}
\end{equation} or $P(|d_j| \leq c ~|~ \mathcal{F}_{j-1})=1$, for $k>2$, $0< c< \infty$. Then for all $x,y>0$, \begin{displaymath}
P\bigg(\sum_{j=1}^n d_j \geq x, V_n^2 \leq y \text{ for some } n \bigg) \leq \exp \Bigg(\frac{-x^2}{2(y+cx)} \Bigg).
\end{displaymath}
\end{lemma} 

\subsection{Proof of Theorem \ref{lem::MartConc}}

Firstly, we demonstrate that a geometric martingale difference sequence meets the conditions of Lemma \ref{lem::delaP}. Define, $Z_i= \sum_{j=1}^i (Y_i-\mu_i)$ and $W_i = Z_i - Z_{i-1}$. By definition $\{Z_i\}_{i=1}^\infty$ is a martingale and $\{W_i\}_{i=2}^\infty$ is a martingale difference sequence. Immediately, from the distribution of $Y_i$, $i \in [n]$, we have $\mathbb{E}(W_i ~|~ \mathcal{F}_{i-1})=0$ and $\mathbb{E}(W_i^2 ~|~ \mathcal{F}_{i-1}) = Var(Y_i ~|~ \mathcal{F}_{i-1}) = \mu_i^2 + \mu_i$. 

The higher-order central moments of the Geometric distribution lack a closed-form expression which makes checking condition \eqref{eq::delapenacondition} more complex. Our technique relies on two main steps: we identify a bound on the \emph{cumulants} of the Geometric distribution, and we use a link between the central moments and cumulants from Combinatorics to realise a central moment bound, given in the following lemma.

\begin{lemma} \label{lem::central_moment_bound}
The central moments $\mu_n$, $n \geq 1$ of the Geometric random variable with parameter $p$ satisfy \begin{displaymath}
\mu_n \leq \frac{!n(1-p)}{p^n}
\end{displaymath} where $!m$ denotes the number of derangements of an integer $m \geq 1$, defined recursively as \begin{displaymath}
!m = (m-1)(!(m-1)+!(m-2))
\end{displaymath} where $!0=1$ and $!1=0$.
\end{lemma} 
The proof of Lemma \ref{lem::central_moment_bound} is given in Section \ref{sec::Lem11}. It uses the property that the the central moments of any distribution may be expressed in terms of the cumulants $\kappa_k$ of the distribution via \emph{incomplete exponential Bell polynomials}. In particular, we have, \begin{equation}
\mathbb{E}(W_i^k ~|~ \mathcal{F}_{i-1}) = \sum_{m=1}^k B_{k,m}(0,\kappa_2,\dots,\kappa_{k-m+1}), \label{eq::comblink}
\end{equation} where the summands $B_{k,m}$ are incomplete exponential Bell polynomials (see e.g. Chapter 11 of \cite{Charalambides2002}). For integers $n \geq m \geq 1$ and arguments $x_1,\dots,x_{n-m+1} \in \mathbb{Z}^{n-m+1}$ these polynomials are defined as \begin{equation}
B_{n,m}(x_1,\dots,x_{n-m+1})= \sum \frac{n!}{j_1! j_2! \dots j_{n-m+1}!}\bigg(\frac{x_1}{1!} \bigg)^{j_1} \bigg(\frac{x_2}{2!} \bigg)^{j_2} \dots \bigg(\frac{x_{n-m+1}}{(n-m+1)!} \bigg)^{j_{n-m+1}}, \label{eq::Bell_polynomial}
\end{equation} where the sum is over all sequences $j_1,j_2,\dots,j_{n-m+1}$ of non-negative integers such that $\sum_{i=1}^{n-m+1} j_i = m$ and $\sum_{i=1}^{n-m+1} i j_i = n$.

The bound in Lemma \ref{lem::central_moment_bound} can be adapted to the form required for condition \eqref{eq::delapenacondition}. We have the following relationship between the number of derangements and the factorial, \begin{displaymath}
!n = \bigg[ \frac{n!}{e} \bigg] \leq \frac{n!}{2}
\end{displaymath} where $[\cdot]$ is the nearest integer function. It follows that \begin{equation}
\mathbb{E}(W_i^k|\mathbb{F}_{i-1}) \leq \frac{k!(1-p)}{2p^k} = \frac{k!}{2} \frac{1-p}{p^2} \frac{1}{p^{k-2}}.
\end{equation} Thus, the central moments of the Geometric random variable $X_i$ satisfy \eqref{eq::delapenacondition} with $\sigma^2=\frac{1-p}{p^2}$ and $c=1/p$.

Thus from Lemma \ref{lem::delaP} we have, for some $n \geq 1$, and any $x > 0$,\begin{displaymath}
P\bigg(\sum_{i=1}^n Y_i - \mu_i \geq x \bigg) \leq \exp\bigg( \frac{-x^2}{2\big(\sum_{i=1}^n \sigma_i^2+ x/(\min_i p_i)\big)}\bigg).
\end{displaymath} Therefore if, for $C>0$, $x=2\log(C)/(\min_i p_i)+\sqrt{2\sum_{i=1}^n \sigma_i^2 \log(C)}$, we have \begin{align*}
P\bigg(\sum_{i=1}^n Y_i - \mu_i \geq x \bigg) &\leq \exp \left( -\frac{  \frac{4\log^2(C)}{(\min_i p_i)^2}+ \frac{4\sqrt{2\sum_{i=1}^n\sigma_i^2\log^{3}(C)}}{\min_i p_i}+2\sum_{i=1}^n \sigma_i^2\log(C)}{\frac{4\log(C)}{(\min_i p_i)^2}+ \frac{2\sqrt{2\sum_{i=1}^n \sigma_i^2 \log(C)}}{\min_i p_i}+2\sum_{i=1}^n \sigma_i^2}\right) \\
&\leq \exp\big(-\log(C) \big) = C^{-1}
\end{align*} By symmetry we have the same bound on $P\big(\sum_{i=1}^n \mu_i - Y_i \geq x \big)$, and the statement of \eqref{eq::MartConcpart1} follows. 

Now consider, \begin{displaymath}
P\Big(2\sum_{i=1}^n Y_i \leq \sum_{i=1}^n \mu_i\Big) = P\Big(\sum_{i=1}^n \mu_i - \sum_{i=1}^n Y_i \geq \frac{\sum_{i=1}^n \mu_i}{2}\Big) \leq P\Big(\sum_{i=1}^n \mu_i - \sum_{i=1}^n Y_i \geq \delta{\sum_{i=1}^n \mu_i}\Big),
\end{displaymath} for any $\delta \in [0,1/2]$. Choosing $$\delta= \left(\frac{2\log(C)}{\min_i p_i}+ \sqrt{2\sum_{i=1}^n \sigma_i^2\log(C)}\right)(\sum_{i=1}^n \mu_i)^{-1},$$ and noting $\delta \leq 1/2$ when $\sum_{i=1}^n \mu_i > 4\log(C)/(\min_i p_i)+ \sqrt{8\sum_{i=1}^n \sigma_i^2\log(C)}$, we have by Lemma \ref{lem::delaP} and a similar manipulation to that used in the proof of \eqref{eq::MartConcpart1}, that $$P\left(2\sum_{i=1}^n Y_i \leq \sum_{i=1}^n \mu_i\right) \leq C^{-1}.$$ Furthermore, since $\sigma_i^2 = \mu_i^2+\mu_i \leq 2\mu_i$ (as $\mu_i\leq 1$), we have also that $$P\left(4\sum_{i=1}^n Y_i \leq \sum_{i=1}^n \sigma_i^2\right) \leq C^{-1}.$$
It follows that \begin{align*}
&\enspace P\bigg(\sum_{i=1}^n Y_i - \mu_i \geq \sqrt{8\sum_{i=1}^n Y_i\log(C)} + \frac{2\log(C)}{\min_i p_i}\bigg) \\
&\leq P\bigg(\sum_{i=1}^n Y_i - \mu_i \geq \sqrt{2\sum_{i=1}^n \sigma_i^2\log(C)} + \frac{2\log(C)}{\min_i p_i} \text{ and } 4\sum_{i=1}^n Y_i \leq \sum_{i=1}^n \sigma_i^2\bigg) \\
&\leq P\bigg(\sum_{i=1}^n Y_i - \mu_i \geq \sqrt{2\sum_{i=1}^n \sigma_i^2\log(Jl^2/2)} + \frac{2\log(C)}{\min_i p_i}\bigg)+ P\bigg( 4\sum_{i=1}^n Y_i \leq \sum_{i=1}^n \sigma_i^2\bigg) \leq 2C^{-1}.
\end{align*} By symmetry we have the high-probability same bound on $\sum_{i=1}^n \mu_i - Y_i $, and the statement of \eqref{eq::MartConcpart2} follows. $\square$

\subsection{Proof of Lemma \ref{lemma::epochUCBconc}} 

We recall that the number of clicks on item $j \in [J]$ in an epoch $l \in [L]$ is a geometric random variable with parameter $p_{j,l}=\sum_{k=1}^K \mathbb{I}\{a_k^l=j\}(1+\lambda_k\alpha_j)^{-1}$, as such $$p_{j,l} \in \left[\left(1+\max_k\lambda_k\alpha_j\right)^{-1},\left(1+\min_k\lambda_k\alpha_j\right)^{-1}\right] \subseteq [0.5,1]$$ for an epoch where $j \in \mathbf{a}^l$. Thus, the sequence of click counts is a sequence of Geometric random variables of the form considered in Theorem \ref{lem::MartConc}. It follows from Theorem \ref{lem::MartConc}, specifically equation \eqref{eq::MartConcpart1}, that for any item $j \in [J]$ the sum of clicks on that item in $L$ epochs obeys, \begin{align*}
 \bigg|\sum_{l=1}^L\sum_{k=1}^K \mathbb{I}\{a_k^l=j\}n_k^l &- \sum_{l=1}^L \sum_{k=1}^K\mathbb{I}\{a_k^l=j\}\lambda_k\alpha_j \bigg| \leq \sqrt{2\sum_{l=1}^L \sigma_{j,l}^2\log\left(JL^2/2\right)} +  4\log\left(JL^2/2\right),
\end{align*} with probability at least $1-4/JL^2$.  As per their definition in equation \eqref{eq::unbiasedestimators} the estimators $\bar{\alpha}_j(l)$ are weighted sums of these click counts, and therefore we have for any $j \in [J]$, $l \in [L]$, and $\mathbf{a}_{1:l}$ such that $\sum_{s=1}^l\sum_{k=1}^K\mathbb{I}\{a_k^s=j\} > 0$, \begin{equation}
\bigg|\bar{\alpha}_{j}(l) - \alpha_j \bigg| \leq \frac{\sqrt{2\sum_{s=1}^l\sigma_{j,s}^2\log(Jl^2/2)}}{\sum_{s=1}^l \sum_{k=1}^K \lambda_k \mathbb{I}\{a_k^s=j\}} + \frac{4\log(Jl^2/2)}{\sum_{s=1}^l \sum_{k=1}^K \lambda_k \mathbb{I}\{a_k^s=j\}},
\end{equation} with probability at least $1-4/Jl^2$. Notice that \begin{align*}
\sigma_{j,l}^2 = \mu_{j,l}^2+\mu_{j,l} \leq 2\mu_{j,l} = 2\alpha_j \sum_{k=1}^K \mathbb{I}\{a_k^l=j\} &\leq 2\sum_{k=1}^K \mathbb{I}\{a_k^l=j\},
\end{align*} and thus we also have, for all $j \in [J]$ and $l \in [L]$, \begin{equation*}
P \left( \left|\bar\alpha_j(l) - \alpha_j \right| > \sqrt{\frac{4\alpha_j\log(Jl^2/2)}{\sum_{s=1}^l \sum_{k=1}^K \lambda_k \mathbb{I}\{a_k^s=j\}}} + \frac{4\log(Jl^2/2)}{\sum_{s=1}^l \sum_{k=1}^K \lambda_k \mathbb{I}\{a_k^s=j\}} ~ \Bigg| ~ \mathbf{a}_{1:l}\right) \leq \frac{4}{Jl^2}.
\end{equation*} Fixing $j$ and $l$ and considering the unconditioned probability, the result stated in equation \eqref{eq::UCBConcpart1} follows via a union bound. 

Similarly, it follows from equation \eqref{eq::MartConcpart2} that the data adaptive bound below holds with probability at least $1-6/JL^2$,\begin{align*}
 \bigg|\sum_{l=1}^L\sum_{k=1}^K \mathbb{I}\{a_k^l=j\}n_k^l &- \sum_{l=1}^L \sum_{k=1}^K\mathbb{I}\{a_k^l=j\}\lambda_k\alpha_j \bigg| \\
&\leq \sqrt{8\sum_{l=1}^L \sum_{k=1}^K \mathbb{I}\{a_k^l=j\}n_k^l\log\left(JL^2/2\right)} +  4\log\left(JL^2/2\right),
\end{align*}  Then by a similar union bound we have the result \eqref{eq::UCBConcpart2} as stated. 

Finally, we consider the probability in equation \eqref{eq::UCBconcpart3}. We have, for $\Lambda_{j,l}$ and $l$ such that $\Lambda_{j,l}>4\log(Jl^2/2)/\alpha_j$, \begin{align*}
P \left(\bar\alpha_j(l) > 2\alpha_j + \frac{4\log(Jl^2/2)}{\Lambda_{j,l}}\right) &\leq P\left( \bar\alpha_j(l)-\alpha_j > \sqrt{\frac{4\alpha_j\log(Jl^2/2)}{\Lambda_{j,l}}} + \frac{4\log(Jl^2/2)}{\Lambda_{j,l}}\right) \leq \frac{2}{Jl^2},
\end{align*} with the final inequality using Lemma \ref{lem::MartConc} once again.
$\square$

\subsection{Proof of Lemma \ref{lem::central_moment_bound}} \label{sec::Lem11}

First, we give a recurrence relation for the cumulants of the Geometric distribution. This will be used to  verify the Bernstein condition for the central moments. These results may also be of independent interest.

\begin{lemma} \label{lem::cumulant_geometric}
The cumulants $\kappa_n$, $n \geq 1$ of the Geometric random variable with parameter $p$ satisfy, \begin{equation}
\kappa_n = \sum_{i=1}^n (-1)^{n-i} \frac{h_{n,i}}{p^i} \label{eq::cumulant_polynomial}
\end{equation} where the coefficients $h_{n,i}$, $n \geq 1$, $i \leq n$ are recursively defined positive integers satisfying \begin{alignat*}{2}
h_{n,1} &= 1 \quad &&\forall n \geq 1 \\
h_{n,i} &= ih_{n-1,i} + (i-1)h_{n-1,i-1} \quad &&\forall n \geq 3, i \in \{2,\dots, n-1\} \\
h_{n,n} &= (n-1)h_{n-1,n-1} &&\forall n \geq 1.
\end{alignat*}
\end{lemma}

\noindent \emph{Proof:} Firstly we note that the cumulants of the Geometric distribution with parameter $p$ satisfy the recurrence relation \begin{equation}
\kappa_k = (p-1) \frac{d\kappa_{k-1}}{dp}, \enspace \kappa_1=\frac{1-p}{p}. \label{eq::differential_recursion}
\end{equation} The second and third cumulants follow immediately from \eqref{eq::differential_recursion} as  \begin{align*}
\kappa_2 &= (p-1)\frac{d\kappa_1}{dp} = (p-1)\bigg( \frac{-1}{p^2}\bigg)= \frac{-1}{p} + \frac{1}{p^2}, \\
\kappa_3 &= (p-1)\frac{d\kappa_2}{dp} = (p-1)\bigg( \frac{1}{p^2} - \frac{2}{p^3}\bigg)= \frac{1}{p} - \frac{3}{p^2} + \frac{2}{p^3}.
\end{align*}
Thus we may verify that $\kappa_3$ satisfies the definition in \eqref{eq::cumulant_polynomial}. Now assume that $\kappa_n$ matches the definition in \eqref{eq::cumulant_polynomial} for some fixed $n>3$ and consider $\kappa_{n+1}$. We have from \eqref{eq::differential_recursion} the following, \begin{align*}
\kappa_{n+1} 	&= (p-1) \frac{d\kappa_n}{dp} \\
				&= (p-1) \sum_{i=1}^n (-1)^{n+1-i} \frac{ih_{n,i}}{p^{i+1}} \\
				&= \sum_{i=1}^n \bigg[(-1)^{n+1-i} \frac{ih_{n,i}}{p^i} -  (-1)^{n+1-i} \frac{ih_{n,i}}{p^{i+1}} \bigg] \\
				&= (-1)^n \frac{h_{n,1}}{p} + \sum_{j=2}^n \bigg[ (-1)^{n+1-j} \frac{jh_{n,j}}{p^j} - (-1)^{n-j} \frac{(j-1)h_{n,j-1}}{p^j}\bigg] - (-1) \frac{n h_{n,n}}{p^{n+1}} \\
				&= (-1)^n \frac{h_{n,1}}{p} + \sum_{j=2}^n \bigg[ \frac{(-1)^{n+1-j}\big(jh_{n,j}+(j-1)h_{n,j-1} \big)}{p^j}\bigg] + \frac{n h_{n,n}}{p^{n+1}},		
\end{align*} thus proving the statement by induction. $\square$

Considering this form of the cumulants \eqref{eq::cumulant_polynomial}, and the nature of the Bell polynomial \eqref{eq::Bell_polynomial}, it is apparent that the $n^{th}$ central moment $\bar\mu_n$ may also be written as $O((1/p)^n)$ polynomials, with some non-negative, integer coefficients $f_{n,1},\dots,f_{n,n}$ (to be specified later). Specifically, we may write \begin{equation}
\bar\mu_n = \sum_{i=1}^n (-1)^{n-i} \frac{f_{n,i}}{p^i}.
\end{equation} 

Next, we introduce a relevant property of a sequence of non-negative integers, and give a lemma showing that the coefficients of the cumulants and central moments have this property. \begin{definition}[Alternating Partial Sum (APS)]
A sequence of $n>0$ non-negative integers, $h_1,\dots,h_n$ is called APS if for all $k \in [n]$ \begin{equation*}
\sum_{i=1}^k (-1)^{n-i} h_i ~ \begin{cases} 
&\geq 0, ~ \text{when } (n-k)\mod 2 =0, \\
&\leq 0, ~ \text{when } (n-k)\mod 2 =1.
\end{cases}
\end{equation*}

\end{definition} 

\begin{lemma} \label{lem::APS_coefficients}
For any integer $n\geq 3$, and Geometric random variable $X$ with parameter $p$, both the coefficients of the polynomial expression for the cumulants of $X$, $h_{n,1},\dots,h_{n,n}$, and the coefficients of the polynomial expression for the central moments of $X$, $f_{n,1},\dots,f_{n,n}$ are APS sequences.
\end{lemma} The full proof of Lemma \ref{lem::APS_coefficients} is in the next subsection. The proof has two main steps. First we show that the sequence $h_{n,1},\dots,h_{n,n}$ is APS for any $n$ from its recursive formula. Second, we show that the sequence $f_{n,1},\dots,f_{n,n}$ can be written as a linear combination of APS sequences (derived from multiplying together cumulants in the Bell polynomial). This linear combination operation preserves the APS property, and thus the sequence $f_{n,1},\dots,f_{n,n}$ is also APS. 

The proof of Lemma \ref{lem::central_moment_bound} follows from application of Lemma \ref{lem::APS_coefficients}. First, we demonstrate that $f_{n,n}=!n$. Recall that the central moments $\mu_n$ are defined in terms of the cumulants as \begin{equation}
\mu_n = \sum_{m=1}^n B_{n,m}(0, \sum_{i=1}^2 (-1)^{2-i} \frac{h_{2,i}}{p^i},\dots, \sum_{i=1}^{n-m+1} (-1)^{n-m+1-i} \frac{h_{n-m+1,i}}{p^i}).
\end{equation} It follows that the leading order coefficient $f_{n,n}$ of the polynomial expression for $\mu_n$ can be expressed in terms of incomplete Bell polynomials of the leading order coefficients of the preceding cumulants, i.e. \begin{equation}
f_{n,n} = \sum_{m=1}^n B_{n,m}(0,h_{2,2},\dots,h_{n-m+1,n-m+1}) = \sum_{m=1}^n B_{n,m}(0,1!,2!,\dots,(n-m)!)
\end{equation} where the second equality follows from Lemma \ref{lem::cumulant_geometric}. The above definition of $f_{n,n}$ coincides with a complete Bell polynomial, such that we have \begin{align}
f_{n,n} &= B_{n}(0,1!,2!,\dots,(n-1)!) \nonumber \\
		&= \sum_{\substack{j_2,\dots,j_n \\ 2j_2+\dots+nj_n=n}} \frac{n!}{j_2!\dots j_n!} \bigg(\frac{1!}{2!}\bigg)^{j_2} \dots \bigg(\frac{(n-1)!}{n!}\bigg)^{j_n} \nonumber \\
		&= \sum_{\substack{j_2,\dots,j_n \\ 2j_2+\dots+nj_n=n}} \frac{n!}{j_2!\dots j_n!} \bigg(\frac{1}{2}\bigg)^{j_2} \dots \bigg(\frac{1}{n}\bigg)^{j_n} = !n. \label{eq::derangement_prop}
\end{align} The final equality follows from the observation that each of the summands in the penultimate expression are the number of permutations in the group of all permutations of $n$ integers with cycle structure $2^{j_2}3^{j_3}\dots n^{j_n}$. By definition this sum is the number of derangements of $n$.

The second stage of the proof is to demonstrate that $\bar\mu_n \leq f_{n,n}/p^n$. First, we note that if $p=1$ then the Geometric variable $X$ has $P(X=0)=1$. Thus, by the definition of $\bar\mu_n$ as a central moment, if $p=1$ then $\bar\mu_n=0$. This implies that the alternating sum of polynomial coefficients $f_{n,1},\dots,f_{n,n}$ must be 0 for any $n$, i.e. \begin{displaymath}
\sum_{i=1}^n (-1)^{n-i} f_{n,i} =0,
\end{displaymath} and in particular, that \begin{equation}
f_{n,n} = \sum_{i=1}^{n-1} (-1)^{n-i} f_{n,i}. \label{eq::nthterm}
\end{equation}

As $p \leq 1$ by definition, the APS property of $f_{n,1},\dots,f_{n-1}$ tells us that \begin{equation}
\sum_{i=1}^{n-1} (-1)^{n-1-i} \frac{f_{n,i}}{p^i} \geq \frac{1}{p^{n-1}}\sum_{i=1}^{n-1} (-1)^{n-1-i}f_{n,i}. \label{eq::property3}
\end{equation} We verify this by considering that \begin{align*}
\sum_{i=1}^{n-1} (-1)^{n-1-i} f_{n,i} &= \sum_{i=1}^{n-1} (-1)^{n-1-i}p^{n-1-i} f_{n,i} + \sum_{i=1}^{n-1-i} (-1)^{n-1-i}(1-p^{n-1-i}) f_{n,i} \\
&\leq \sum_{i=1}^{n-1} (-1)^{n-1-i}p^{n-1-i} f_{n,i},
\end{align*} where the inequality holds since the second sum is negative. Dividing both sides by $p^{n-1}$ gives \eqref{eq::property3}.

We then complete the proof by bounding $\bar\mu_n$ as follows \begin{align*}
\bar\mu_n 	= \sum_{i=1}^n (-1)^{n-i} \frac{f_{n,i}}{p^i}
			&= \frac{f_{n,n}}{p^n} + \sum_{i=1}^{n-1} (-1)^{n-i} \frac{f_{n,i}}{p^i} \\ 
			&\leq \frac{f_{n,n}}{p^n} - \frac{f_{n,n}}{p^{n-1}} \\
			&= \frac{f_{n,n}(1-p)}{p^n} = \frac{!n (1-p)}{p^n},
\end{align*} where the inequality follows from \eqref{eq::property3} and  \eqref{eq::nthterm}, and the final equality follows from \eqref{eq::derangement_prop}. $\square$

\subsection{Proof of Lemma \ref{lem::APS_coefficients}}
Firstly we show by induction that the sequences $h_{n,1},\dots,h_{n,n}$ are APS for all $n \geq 3$. 

Consider first the case of $n=3$. We have, as defined in Lemma \ref{lem::cumulant_geometric}, that $h_{3,1}=1$, $h_{3,1}-h_{3,2} = 1-3=-2$, and $h_{3,1}-h_{3,2}+h_{3,3}=1-3+2=0$. Thus all of the non-negativity and non-positivity conditions are satisfied and the sequence $h_{3,1},h_{3,2},h_{3,3}$ is APS. We now assume for some fixed $m \geq 4$ that $h_{m,1},\dots,h_{m,m}$ is APS, and proceed to consider the sequence $h_{m+1,1},\dots,h_{m+1,m+1}$.

By definition we have $h_{m,1}=1$ and $h_{m+1,m+1}=m!$. Thus the APS conditions are satisfied for $k=1$ and $k=m+1$. We proceed to consider $\sum_{i=1}^k (-1)^{m+1-i} h_{m+1,k}$ for $k \in \{2,\dots,m\}$. We have, \begin{align}
\sum_{i=1}^k (-1)^{m+1-i} h_{m+1,i} &= (-1)^{m}h_{m,1} + \sum_{i=2}^k (-1)^{m+1-i} \big[ih_{m,i}+(i-1)h_{m,i-1} \big] \nonumber \\
&= (-1)^{m+1-k} kh_{m,k}. \label{eq::partial_sum}
\end{align} Since all $h_{m,k}$, $m \geq 2, k \leq m$ are positive integers, \eqref{eq::partial_sum} is positive when $m+1-k \mod 2=0$ and negative when $m+1-k \mod 2=1$. Thus the APS conditions are satisfied for the sequence $h_{m+1,1}, \dots,h_{m+1,m+1}$ given $h_{m,1},\dots,h_{m,m}$ is APS. Thus, by induction, the sequences $h_{n,1},\dots,h_{n,n}$ are APS for all $n \geq 3$.

We next show two properties of APS sequences. Firstly, we have the property that addition preserves APS.
\begin{property}[Preservation of APS under addition]
If $a_{1},\dots,a_{n}$ and $b_{1},\dots,b_{n}$ are APS sequences, then the sequence  $a_{1}+b_{1},\dots,a_{n}+b_{n}$  is APS.
\end{property}
To verify, consider first $j \leq n: n-j \mod 2=0$, we have \begin{displaymath}
\sum_{i=1}^j (-1)^{n-i} (a_{i}+b_{i}) = \sum_{i=1}^j (-1)^{n-i}a_{i} + \sum_{i=1}^j (-1)^{n-i} b_{i} \geq 0,
\end{displaymath} since $a_{1},\dots,a_{n}$ and $b_{1},\dots,b_{n}$ are both APS. Similarly, for $j  \leq n: n-j \mod 2=1$ we have \begin{displaymath}
\sum_{i=1}^j (-1)^{n-i} (a_{i}+b_{i}) = \sum_{i=1}^j (-1)^{n-i}a_{i} + \sum_{i=1}^j (-1)^{n-i} b_{i} \leq 0,
\end{displaymath} showing that $(a_{1}+b_{1},\dots,a_{n}+b_{n})$ are APS.

The second property states that if two polynomials (in the same variable) have APS coefficients, the product of these polynomials has APS coefficients. \begin{property}[Preservation of APS under polynomial multiplication] Let $a_1,\dots,a_n$ and $b_1,\dots,b_m$ be APS for $n,m \in \mathbb{N}$, with $\sum_{i=1}^n (-1)^{n-i}a_i=0$ and $\sum_{i=1}^m (-1)^{m-i}b_i=0$. Then, the sequence of polynomial coefficients  $c_1,\dots, c_{n+m}$ such that \begin{displaymath}
\sum_{i=1}^{n+m} (-1)^{n+m-i} c_ix^i = \bigg(\sum_{i=1}^n (-1)^{n-i} a_ix^i \bigg)\bigg(\sum_{i=1}^m (-1)^{m-i} b_ix^i \bigg), \quad x \in \mathbb{R}
\end{displaymath} are APS.

\end{property} To verify this, consider \begin{align*}
\sum_{i=1}^{n+m} (-1)^{n+m-i}c_i x^i &= \big(a_nx^n -a_{n-1}x^{n-1} + \dots + (-1)^{n-1} a_1x\big) \sum_{j=1}^m (-1)^{m-j}b_jx_j \\
&= \sum_{j=1}^m (-1)^{m-j} b_j \big(a_nx^{n+j} - a_{n-1}x^{n-1+j} + \dots + (-1)^{n-1}a_1x^{1+j} \big) \\
&=\sum_{j=1}^m \sum_{i=1}^{n+m} (-1)^{n+m-i} d_i^{(j)} x^i
\end{align*} for coefficients $d_i^{(j)}$, $j \in [m], i \in [n+m]$, defined as follows \begin{align*}
d_i^{(j)} = \begin{cases} 
&0, \quad \quad \quad i < 1+j \\
&a_{i-j}b_j, \quad i \in \{1+j,\dots,n+j\} \\
&0, \quad \quad \quad i > n+j
\end{cases}
\end{align*} Now, since $a_1,\dots,a_n$ is APS and has $\sum_{i=1}^n (-1)^{n-i}a_i=0$, we have \begin{align*}
\sum_{i=1}^k (-1)^{n+m-i}d_i^{(j)} = \begin{cases}
&0, \quad \quad \quad k < 1+j \\
&\sum_{i=1}^{k-j} (-1)^{n-i} a_i, \quad k \in \{1+j,\dots,n+j\} \\
&0, \quad \quad \quad k > n+j
\end{cases}
\end{align*} for all $j \in [m]$. Thus the sequence $\{d_i^{(j)}\}_{i=1}^{n+m}$ is APS for each $j \in [m]$ and by Property 1, the coefficients $c_1,\dots,c_{n+m}$ are also APS.

Using the result that $h_{n,1},\dots,h_{n,n}$ is APS for any $n \geq 3$ and the above properties, we will show that the sequences $f_{n,1},\dots,f_{n,n}$ are also APS for all $n \geq 3$. Firstly, we recall the definition of the central moments in terms of cumulants, \begin{displaymath}
\bar\mu_n = \sum_{k=1}^n \sum_{\substack{j_2,\dots,j_n: \\ j_2+\dots j_n=k \\ 2j_2+\dots+nj_n=n}} \frac{n!}{j_2!\dots j_n!} \bigg(\frac{\kappa_2}{2!} \bigg)^{j_2} \bigg(\frac{\kappa_3}{3!} \bigg)^{j_3} \dots \bigg(\frac{\kappa_n}{n!} \bigg)^{j_n}.
\end{displaymath} Since $\kappa_2=O(p^{-2})$, $\kappa_3=O(p^{-3})$, and $\kappa_n=O(p^{-n})$, each summand is $o(p^{-n})$, and may be written as a polynomial \begin{displaymath}
\sum_{i=1}^n (-1)^{n-i} \frac{J_i}{p^i} = \frac{n!}{j_2!\dots j_n!} \bigg(\frac{\kappa_2}{2!} \bigg)^{j_2} \bigg(\frac{\kappa_3}{3!} \bigg)^{j_3} \dots \bigg(\frac{\kappa_n}{n!} \bigg)^{j_n}
\end{displaymath} where $J_1,\dots,J_n$ are (possibly zero or negative) coefficients which are a function of variables $j_2,\dots,j_n$, the order $n$ of the moment in question, and the sequences $h_{m,1},\dots,h_{m,n}$ for all $m \leq n$. It follows by Property 2 that the coefficients $J_1,\dots J_n$ are APS. Finally since the coefficients $f_{n,1},\dots,f_{n,n}$ can be written as sums of the $J$ coefficients over the valid combinations of $j_2,\dots,j_n$, we have by Property 1 that the coefficients $f_{n,1},\dots,f_{n,n}$ are APS. $\square$

\section{Proof of Maximum Likelihood Concentration} \label{app::MLconc}

In this section we provide proofs of Theorem \ref{thm::MVgeneral} and Corollary \ref{corr:MLEconc} giving our result on the concentration of the maximum likelihood estimators in the unknown position bias setting.

\subsection{Proof of Theorem \ref{thm::MVgeneral} }

The first step of the proof is derive a bound on the relative entropy of functions with respect to the product measure. We make use of the following result from \cite{JoulinPrivault2004}. It gives a logarithmic Sobolev inequality tailored to functions of a univariate Geometric distribution. 
\begin{theorem}[Theorem 3.7 \citep{JoulinPrivault2004}] \label{thm::JoulinPrivault}
Let $\pi$ denote the law of a Geometric random variable with parameter $p$. Let $0<b<-\log(1-p)$ and let $f: \mathbb{N} \rightarrow \mathbb{R}$ such that $|d^+f|=\max_{k \in \mathbb{N}}|f(k+1)-f(k)|\leq b$ for all $k \in \mathbb{N}$. We have \begin{equation*}
Ent_\pi\left(e^f\right) \leq \frac{(1-p)e^b}{p(1-\sqrt{(1-p)e^b})}\mathbb{E}_\pi\left(|d^+f|^2e^f\right).
\end{equation*}
\end{theorem} 

The chain rule for differential entropy (see e.g. Theorem 8.6.2 of \cite{CoverThomas2012}) states that for a series of random variables $X_1,\dots,X_n$ with joint distribution $\mu^n$, \begin{displaymath}
Ent_{\mu^n}\left(X_1,\dots,X_n\right) =  \sum_{l=1}^n Ent_{\mu_l}\left(X_l ~|~ X_1,\dots,X_{l-1}\right).
\end{displaymath} Using this result we may extend the univariate bound in Theorem \ref{thm::JoulinPrivault} in both dimensions to a bound under product measure. Specially, for every function $G: \mathbb{N}^{d \times n} \rightarrow \mathbb{R}$, satisfying $DG_{li} \leq b_{li}$ for constants $0< b_{li} < -\log(1-p_{li})$ $i \in [d]$, $l \in [n]$, we have \begin{equation}
Ent_{\mu^n} \left(e^G\right) \leq \max_{i \in [d], l \in [n]} \frac{(1-p_{li})e^{b_{li}}}{p_{li}(1-\sqrt{(1-p_{li})e^{b_{li}})}} \mathbb{E}_{\mu^n} \left(\sum_{i=1}^{d} \sum_{l=1}^n |d^+ G_{li}|^2 e^{G}\right). \label{eq::tensorisedentropy}
\end{equation} In particular, under the assumptions of the theorem statement, this holds for $F$, with $\max_{l,i} b_{li}=\beta_2$.

We now follow the so-called Herbst's method \citep{DaviesSimon1984,AidaEtAl1994} to achieve a deviation inequality on $F$. For ease in what follows we introduce the function $M_G: \mathbb{R} \rightarrow \mathbb{R}$ with \begin{equation}
M_G(b) = \max_{i \in [d],l \in [n]} \frac{(1-p_{li})e^b}{p_{li}(1-\sqrt{(1-p_{li})e^b})},  \quad b \in \left(0, \max_{i \in [d],l\in [n]} -\log(1-p_{li})\right). \label{eq::MGdef}
\end{equation} Further we let $\bar{p}$ be the parameter attaining the maximum in \eqref{eq::MGdef}, i.e. $$\bar{p}=\argmax_{p_{li}: i \in{d},l\in[n]} \frac{(1-p_{li})e^b}{p_{li}(1-\sqrt{(1-p_{li})e^b})},$$ for any valid $b$.

Applying \eqref{eq::tensorisedentropy} to $\eta F$ for every $0 < \eta \leq -\log(1-\bar{p})/(2\beta_2)$ and substituting the definition for entropy we have, \begin{align*}
&\mathbb{E}_{\mu^n}\left(\eta F e^{\eta F}\right) - \mathbb{E}_{\mu^n} \left(e^{\eta F}\right)\log\left(\mathbb{E}_{\mu^n} \left(e^{\eta F}\right)\right) \leq M_G(\eta\beta_2) \mathbb{E}_{\mu^n}\left(\sum_{l=1}^n \sum_{i=1}^{d_1}\eta^2|F(\mathbf{X} + \epsilon_{li}) - F(\mathbf{X})|^2 e^{ \eta F}\right).
\end{align*} Exploiting the assumed bound \eqref{eq::quadraticFDGbound}, and introducing the notation $H(\eta)=\mathbb{E}_{\mu^n}\left(e^{\eta F}\right)$ we may then rewrite the above display as, \begin{align*}
\eta H'(\eta) - H(\eta)\log\left(H(\eta)\right) \leq \eta^2 \beta_1^2 M_G(\eta\beta_2) H(\eta).
\end{align*}  We then set $K(\eta) = \frac{1}{\eta}\log(H(\eta))$, with $K(0)=H'(0)/H(0)=\mathbb{E}_{\mu^n}(F)$ and observe, \begin{align*}
K'(\eta) \leq \frac{\eta^2\beta_1^2M_G(\eta\beta_2)H(\eta)}{\eta^2 H(\eta)} = \beta_1^2M_G(\eta\beta_2) \leq \beta_1^2M_G\left(\frac{-\log(1-\bar{p})}{2}\right)
\end{align*} since $M_G$ is an increasing function. We will define $M:= M_G\left(\frac{-log(1-\bar{p})}{2}\right)$ for convenience. As such we may bound $K(\eta) \leq \mathbb{E}_{\mu^n}(F) + \eta\beta_1^2 M$, and derive the exponential inequality, \begin{equation}
\mathbb{E}_{\mu^n}\left(e^{\eta F}\right) \leq \exp\left(\eta\mathbb{E}_{\mu^n}\left(F\right) + \eta^2\beta_1^2M \right), \quad 0 < \eta \leq \frac{-\log(1-\bar{p})}{2\beta_2}. \label{eq::exp.geom}
\end{equation} 

Finally, we apply a Chernoff bound to $F$, and substitute \eqref{eq::exp.geom}, giving, \begin{align*}
\mathbb{P}_{\mu^n}\left(F \geq \mathbb{E}_{\mu^n}(F)+\delta \right) &\leq \exp \left(\eta\mathbb{E}_{\mu^n}\left(F \right) + \eta^2\beta_1^2M - \eta\mathbb{E}_{\mu^n}(F) - \eta\delta \right)
\end{align*} which when minimised over $\eta \in (0,-\log(1-\bar{p})/(2\beta_2)]$, yields the stated result. $\square$

\subsection{Proof of Corollary \ref{corr:MLEconc}}

The function $\alpha^{EM}$ which computes the EM estimates of $\bs\alpha$, is posed as a function from $\mathbb{N}^{K \times J} \times \mathbb{N}^{K \times J}$ to $[0,1]^J$, where the input matrices are of the form of $\mathbf{N}(L)$ and $\tilde{\mathbf{N}}(L)$. Recall that we have $\alpha^{EM}(\mathbf{N}(L),\tilde{\mathbf{N}}(L))=\bs\alpha^{EM}$ where $\bs\alpha^{EM}$ are the EM estimates of $\bs\alpha$ derived from Algorithm \ref{alg::EMinf}. For the purposes of this proof we will define $\bar\alpha^{EM}: \mathbb{N}^{K\times L}\times [J]^{K \times L} \rightarrow [0,1]^J$, which computes the same $\bs\alpha^{EM}$ estimates but via an alternative arrangement of the input data. 

Let $\bar{\mathbf{N}}(L)$ be the $K \times L$ matrix whose $l^{th}$ column ($l \in [L]$) is $\mathbf{n}^l$, which we recall is the vector of clicks per slot in epoch $l$.  Let $\bar{\mathbf{A}}(L)$ be the $K \times L$ matrix whose $l^{th}$ column ($l \in [L]$) is $\mathbf{a}^l$ the action vector in epoch $l$. Define $\bar\alpha^{EM}$ such that $\bar\alpha^{EM}(\bar{\mathbf{N}}(L),\bar{\mathbf{A}}(L)) = \alpha^{EM}(\mathbf{N}(L),\tilde{\mathbf{N}}(L))=\bs\alpha^{EM}$.

Then, for fixed $\bar{\mathbf{A}}(L)$, the restriction of $\bar\alpha^{EM}$ to its $j^{th}$ output, $\bar\alpha_j^{EM}$, is, a function from $\mathbb{N}^{K \times L}$ to $[0,1]$. It also follows from the definitions above that $\mathbb{E}(\bar\alpha_j^{EM})=\alpha_j$. Finally since the entries of the (non-fixed) input matrix $\bar{\mathbf{N}}(L)$ are Geometric random variables, the function $\bar\alpha_j^{EM}$ fits within the framework of Theorem \ref{thm::MVgeneral}. We therefore have that\begin{displaymath}
P\left(|\alpha_j^{EM}(\mathbf{N}(L),\tilde{\mathbf{N}}(L))-\alpha_j| \geq \delta \right) = P\left(|\bar\alpha_j^{EM}(\bar{\mathbf{N}}(L),\bar{\mathbf{A}}(L))-\alpha_j| \geq \delta\right) \leq \exp \left\lbrace \frac{-\delta^2}{4\beta_{1,L,j}^2 M}\right\rbrace
\end{displaymath} where \begin{equation}
\beta_{1,j,l} =\sup_{\mathbf{X} \in \mathbb{N}^{K \times L}} \sqrt{\sum_{l=1}^L \sum_{k=1}^K \left|\bar\alpha_{j}^{EM}(\mathbf{X}+ \epsilon_{lk})-\bar\alpha_{j}^{EM}(\mathbf{X}) \right|^2} \label{eq::beta1jl}
\end{equation} for any $\mathbf{X} \in \mathbb{N}^{K \times L}$.

We recall that $M= M_G(\max_{k \in [K], l \in [L]} - \log(1-p_{kl})/2)$. Specifically in the context of our MNL-LTR problem, we have $p_{kl} \in [1/2,1)$ for all $k \in [K]$, $l \in [L]$. Thus, $M \leq \sqrt{0.5}/(0.5(1-(0.5)^{1/4})) \leq 9$. Rearranging the exponential inequality and substituting a bound on $M$ we therefore have, \begin{displaymath}
P\left(|\alpha_j^{EM}(\mathbf{N}(L),\tilde{\mathbf{N}}(L)) - \alpha_j| \geq \sqrt{72\beta_{1,L,j}^2\log(\sqrt{J}L)} \right) \leq \frac{2}{JL^2}. \quad \square
\end{displaymath}

\subsection{Proof of Lemma \ref{lem::betabound}}

In this section, we derive the bound on the sensitivity parameter $\beta_{1,j,l}$ which enables an analysis of the regret. We first derive an alternative expression for $\beta_{1,j,l}$ to that in \eqref{eq::beta1jl} in terms of the $\alpha^{EM}$ functions, as oppose to the $\bar\alpha^{EM}$ functions, as \begin{displaymath}
\beta_{1,j,l} = \sup_{\mathbf{N}\in\mathbb{N}^{K \times J}} \sqrt{\sum_{i=1}^J \sum_{k=1}^K \tilde{N}_{ki}\left|\alpha_j^{EM}(\mathbf{N},\tilde{\mathbf{N}})-\alpha_j^{EM}(\mathbf{N}+\epsilon_{ki},\tilde{\mathbf{N}})\right|^2}.
\end{displaymath} Bounding the finite differences with partial derivatives, we have the following bound on the sensitivity parameter, \begin{equation}
\beta_{1,j,l} \leq \sup_{\mathbf{N}\in\mathbb{N}^{K \times J}} \sqrt{\sum_{i=1}^J \sum_{k=1}^K \tilde{N}_{ki} \max_{n_{ki} \in [N_{ki},N_{ki}+1]}\left(\frac{\partial \alpha_j^{EM}}{\partial N_{ki}}\bigg|_{N_{ki}=n_{ki}}\right)^2}. \label{eq::betabound}
\end{equation} In what follows we shall obtain a further analytical bound on this expression in order to relate the sensitivity parameter to the selected actions and number of rounds.

For the purposes of this section, let $\bs\theta=(\bs\alpha,\bs\lambda_{-1})$ denote the length $J+K-1$ vector of the unknown parameters in the unknown position biases model. As before, $\mathbf{N}$ and $\tilde{\mathbf{N}}$ are $J\times K$ matrices of click- and play-counts for each item-slot combination. We will suppress dependence on $L$ for brevity in this section. In order to express gradients with respect to click counts conveniently, we introduce the notation $\mathbf{n}$ to represent a vectorised version of $\mathbf{N}$. Here element $m$ of $\mathbf{n}$ corresponds to element $N_{(m \mod K), \lceil m/K \rceil}$ of the matrix $\mathbf{N}$. 

For a given pair $\mathbf{n},\tilde{\mathbf{N}}$, the maximum likelihood estimate, $$\bs\theta^*(\mathbf{n},\tilde{\mathbf{N}}) \in \argmax_{\bs\theta \in [0,1]^{J+K-1}} \ell\left(\bs\theta; \mathbf{n},\tilde{\mathbf{N}}\right),$$ is found where $\nabla_{\bs\theta}\ell(\bs\theta;\mathbf{n},\tilde{\mathbf{N}})=0$. It then follows from the continuous differentiability of the likelihood, and an application of the Implicit Function theorem that the gradient of the maximiser with respect to the click vector $\mathbf{n}$, within an open neighbourhood of a particular solution $\bs\theta^*_0$ may be written, \begin{equation}
\nabla_{\mathbf{n}}\bs\theta^*(\mathbf{n},\tilde{\mathbf{N}}) = -\left[\nabla_{\bs\theta\bs\theta}\ell\left(\bs\theta;\mathbf{n},\tilde{\mathbf{N}}\right) \right]^{-1}\nabla_{\bs\theta\mathbf{n}}\ell\left(\bs\theta;\mathbf{n},\tilde{\mathbf{N}}\right). \label{eq::gradbyimpfunc}
\end{equation}  Elements of $\nabla_{\mathbf{n}}\bs\theta^*$, a $(J+K-1)\times JK$ matrix provide the gradients of maximum likelihood estimators contained in \eqref{eq::betabound}.

To compute \eqref{eq::gradbyimpfunc}, we first consider the gradient vector $\nabla_{\bs\theta}\ell$ which contains the partial derivatives, \begin{align*}
\frac{\partial\ell}{\partial\alpha_j}=\sum_{k=1}^K \frac{N_{kj}}{\alpha_j} - \frac{\lambda_k(N_{kj}+\tilde{N}_{kj})}{(1+\alpha_j\lambda_k)} ~j\in[J], \quad 
\frac{\partial\ell}{\partial\lambda_{k}}=\sum_{j=1}^J \frac{N_{kj}}{\lambda_k} - \frac{\alpha_j(N_{kj}+\tilde{N}_{kj})}{(1+\alpha_j\lambda_k)}~ k\in[K]\setminus \{1\}.
\end{align*} Thus, the second derivatives contained in the Hessian matrix $\nabla_{\bs\theta\bs\theta}\ell$ are of the following form \begin{align*}
\frac{\partial^2\ell}{\partial\alpha_j^2}= \sum_{k=1}^K  \frac{\lambda_k^2(N_{kj}+\tilde{N}_{kj})}{(1+\alpha_j\lambda_k)^2} -\frac{N_{kj}}{\alpha_j^2}, \quad 
&\frac{\partial^2\ell}{\partial\lambda_k^2}= \sum_{j=1}^J  \frac{\alpha_j^2(N_{kj}+\tilde{N}_{kj})}{(1+\alpha_j\lambda_{k})^2} - \frac{N_{kj}}{\lambda_{k}^2},  \quad \frac{\partial^2\ell}{\partial\alpha_j\lambda_k}= -\frac{(N_{kj}+\tilde{N}_{kj})}{(1+\alpha_j\lambda_{kj})^2},
\end{align*} with $\partial^2\ell/\partial\alpha_j\partial\alpha_j'=0$ where $j\neq j'$  and $\partial^2\ell/\partial\lambda_k\partial\lambda_k'=0$ where $k\neq k'$. The Hessian therefore has a sparsity structure which we later exploit: \begin{align*}
    \nabla_{\bs\theta\bs\theta}\ell = \left(\begin{array}{*8{C{3em}}}
        \frac{\partial^2 \ell}{\partial \alpha_1^2} & 0 &  \dots &  0 & \frac{\partial^2 \ell}{\partial \alpha_1 \partial \lambda_2} & \dots & \dots&\frac{\partial^2 \ell}{\partial \alpha_1 \partial \lambda_K} \\
        0 & \ddots  &  & \vdots &\vdots & & & \vdots \\
        \vdots &    & \ddots & \vdots & \vdots &  & &  \vdots \\
        0 &   \dots  & 0 & \frac{\partial^2 \ell }{\partial \alpha_J^2} & \frac{\partial^2 \ell}{\partial \alpha_J \partial \lambda_2} & \dots & \dots &\frac{\partial^2 \ell}{\partial \alpha_J \partial \lambda_K} \\
        \frac{\partial^2 \ell}{\partial \alpha_1 \partial \lambda_2} &  \dots  & \dots & \frac{\partial^2 \ell}{\partial \alpha_J \partial \lambda_2} & \frac{\partial^2 \ell}{\partial \lambda_2^2} & 0 & \dots & 0 \\
        \vdots & &  & \vdots & 0 &\ddots & & \vdots \\
        \vdots & &  & \vdots & \vdots  & & \ddots & 0 \\
        \frac{\partial^2 \ell}{\partial \alpha_1 \partial \lambda_K} &  \dots & \dots & \frac{\partial^2 \ell}{\partial \alpha_J \partial \lambda_K} & 0 & \dots  & 0 & \frac{\partial^2 \ell}{\partial \lambda_K^2}
    \end{array}\right).
\end{align*} Finally, the elements of $\nabla_{\bs\theta\mathbf{n}}\ell$ are the second derivatives, \begin{equation}
\frac{\partial^2\ell}{\partial\alpha_j \partial N_{kj}}= \frac{1}{\alpha_j} - \frac{\lambda_k}{(1+\alpha_j\lambda_k)}, \quad \frac{\partial^2\ell}{\partial\lambda_k \partial N_{kj}}= \frac{1}{\lambda_k}-\frac{\alpha_j}{(1+\alpha_j\lambda_k)},\label{eq:theta_n}
\end{equation} and $\partial^2\ell/\partial\alpha_jN_{kj'}=0$ for $j\neq j'$ and $\partial^2\ell/\lambda_kN_{k'j}=0$ for $k\neq k'$. 

It follows from the sparsity of $\nabla_{\bs\theta\mathbf{n}}\ell$ that the gradient of a particular attractiveness parameter estimate $\alpha_j^*$ with respect to a particular click count $N_{kl}$, $k \in [K]$, $l\in[J]$, may be written, \begin{align}
\frac{\partial\alpha_j^*}{\partial N_{kl}}&=-\sum_{s=1}^J \left(\nabla_{\bs\theta\bs\theta}^{-1}\right)_{js}\frac{\partial^2\ell}{\partial\alpha_s\partial N_{kl}}- \sum_{m=2}^K \left(\nabla_{\bs\theta\bs\theta}^{-1}\right)_{j,J+m-1}\frac{\partial^2\ell}{\partial\lambda_m\partial N_{kl}} \nonumber \\
&= -\left(\nabla_{\bs\theta\bs\theta}^{-1}\right)_{jl}\frac{\partial^2\ell}{\partial\alpha_l\partial N_{kl}} - \mathbb{I}\{k\neq 1\}\left(\nabla_{\bs\theta\bs\theta}^{-1}\right)_{j,J+k-1}\frac{\partial^2\ell}{\partial\lambda_k\partial N_{kl}}, \label{eq::dalphaNk}
\end{align} where $\nabla_{\bs\theta\bs\theta}^{-1}$ will henceforth be shorthand for the inverse Hessian, $(\nabla_{\bs\theta\bs\theta}\ell)^{-1}$. 
We proceed to bound these gradients uniformly, and apply said bounds to derive a bound on $\beta_{1,j,l}$. 

To bound the elements of $\nabla_{\bs\theta\bs\theta}^{-1}$, recall the definition of the matrix inverse as $A^{-1}=(\det(A))^{-1}\text{adj}(A)$, where $\text{adj}(A)$ denotes the adjugate of a square matrix $A$, a matrix of the same dimension, consisting of minors of the matrix. Specifically, entry $(h,i)$ of $\text{adj}(A)$ is the determinant of the matrix obtained by removing the $i^{th}$ row and $h^{th}$ column of $A$, times $(-1)^{h+i}$, i.e. $\text{adj}(A)_{hi}=(-1)^{h+i}\det(A_{-i,-h})$. We may therefore write, for $j\in[J]$, $k\in[K]$, and $l\in[J]$, \begin{align*}
    \frac{\partial \alpha_j^*}{\partial N_{kl}} &= - \frac{\left(\text{adj}(\nabla_{\bs\theta\bs\theta}) \right)_{jl}}{\det(\nabla_{\bs\theta\bs\theta})} \frac{\partial^2 \ell}{\partial \alpha_l \partial N_{kl}} - \mathbb{I}\{k\neq 1\} \frac{\left(\text{adj}(\nabla_{\bs\theta\bs\theta}) \right)_{j,J+k-1}}{\det(\nabla_{\bs\theta\bs\theta})} \frac{\partial^2 \ell}{\partial \lambda_k\partial N_{kl}},
\end{align*} and, since the second derivatives which mix parameter and click derivatives (those of the form $\partial^2\ell/\partial\alpha_\cdot \partial N_{\cdot\cdot}$ and $\partial^2\ell/\partial\lambda_\cdot \partial N_{\cdot\cdot}$ defined in \eqref{eq:theta_n}) are independent of $\mathbf{N}$ and $\tilde{\mathbf{N}}$, there exists a constant $C_1>0$ such that,  \begin{align}
    \beta_{1,j,L} &= \sqrt{\sum_{l=1}^J\sum_{k=1}^K \tilde{N}_{kl}\max_{n}\frac{\partial\alpha_j}{\partial N_{kl}}^2} \nonumber \\
    &\leq \sqrt{C_1\sum_{l=1}^J \left(\tilde{N}_{1l}\left(-\frac{(\text{adj}(\nabla_{\bs\theta\bs\theta}))_{jl}}{\det(\nabla_{\bs\theta\bs\theta})}\right)^2 + \sum_{k=2}^K\tilde{N}_{kl}\left(-\frac{(\text{adj}(\nabla_{\bs\theta\bs\theta}))_{jl}}{\det(\nabla_{\bs\theta\bs\theta})}- \frac{(\text{adj}(\nabla_{\bs\theta\bs\theta}))_{j,J+k+1}}{\det(\nabla_{\bs\theta\bs\theta})}\right)^2\right)}\nonumber \\
    &= \sqrt{\frac{C_1\sum_{l=1}^J \tilde{N}_{1l}\left(\text{adj}(\nabla_{\bs\theta\bs\theta})_{jl}\right)^2 + \sum_{k=2}^K \tilde{N}_{kl}\left(\text{adj}(\nabla_{\bs\theta\bs\theta})_{jl}+\text{adj}(\nabla_{\bs\theta\bs\theta})_{j,J+k-1}\right)^2}{\det(\nabla_{\bs\theta\bs\theta})^2}}. \label{eq::beta_adj_det}
\end{align}

For any $j\in [J]$, $k\in[K]\setminus\{1\}$, the variable $\tilde{N}_{kj}$ enters (always linearly) in four elements of $\nabla_{\bs\theta\bs\theta}\ell$ only, namely those indexed $(j,j), (j,J+k-1), (J+k-1,j),$ and $(J+k-1,J+k-1)$. The determinant computation never takes a product of terms in the same row or column, and thus, \begin{align*}
    \det (\nabla_{\bs\theta\bs\theta}\ell) = O\left(\tilde{N}_{kj}^2 \right), ~~~ \forall j \in [J], ~ k\in [K]\setminus \{1\}.
\end{align*} For the adjugate of our Hessian matrix, we have the following order results, for $h,i\in[J+K-1]$, $j\in[J]$, and $k\in[K]$, \begin{align*}
    \left|\left(\text{adj}(\nabla_{\bs\theta\bs\theta})\right)_{hi}\right| &=\begin{cases}  & O\left(\tilde{N}_{kj}\right) \quad \text{ when } h \in \{j, J+k-1\} \text{ and/or } i  \in \{j, J+k-1\}, \\
    & O\left(\tilde{N}_{kj}^2 \right) \quad \text{ otherwise}.
    \end{cases}
\end{align*} That is, the elements of the adjuagte (themselves determinants of submatrices), are quadratic in variables $\tilde{N}_{kj}$ unless they are contained within a row or column sharing indices of the variable. In such rows and columns, the order is linear, since instances of  $\tilde{N}_{kj}$ are removed from the determinant computation. The order never falls to $O(1)$ because the removal of no single row $h$ and column $i$ can remove all four instances of the variable $\tilde{N}_{kj}$. 

It follows, from these order results that within the square root of Equation \eqref{eq::beta_adj_det}, for all $k\in[K]$ that the numerator is of order $O\left(\tilde{N}_{kj}^3\right)$, while the denominator is $O\left(\tilde{N}_{kj}^4\right)$ and thus there exists a constant $C>0$ such that, \begin{align*}
\beta_{1,j,L} \leq C \sqrt{\frac{JK}{\max_k \tilde{N}_{kj}}}~. ~~~~~\square
\end{align*}

\section{Proof of Regret Upper Bounds} \label{app::regret}

In this section we provide a proof of Proposition \ref{thm::PBregret}, giving an upper bound on the regret of the Epoch-UCB algorithm for the known position bias setting. The proof has two main stages. First we construct an event that all the UCB indices remain in intervals of certain width around the unknown attractiveness parameters, and verify that this is a high-probability event. We simply assume that the regret is the worst possible if this event does not occur. Then conditioned on the high probability event occurring, we utilise bounds on the values of the UCB indices and the properties of the reward function to bound the regret per epoch, which is aggregated over $L$ epochs to give the stated regret bound.

\subsection{Proof of Proposition \ref{thm::PBregret}}
We begin by defining the regret specifically for an epoch-based algorithm. We have that the $T$-round regret as defined in \eqref{eq::regret} may be rewritten as \begin{displaymath}
Reg(T) = \mathbb{E}\left(\sum_{l=1}^L \sum_{t\in \mathcal{E}_l} R(\mathbf{a}^*)-R\left(\mathbf{a}^l\right) \right) = \mathbb{E}\left(\sum_{l=1}^L |\mathcal{E}_l| \left(R(\mathbf{a}^*)-R\left(\mathbf{a}^l\right)\right) \right).
\end{displaymath} We recall that $|\mathcal{E}_l|$, the number of rounds in epoch $l$ follows a Geometric distribution when conditioned on $\mathbf{a}^l$, and that $\mathbf{a}^l$ is a deterministic function of the history $\mathcal{H}_{l-1}$. As such we may use the the law of conditional expectations to replace $|\mathcal{E}_l|$ with its expectation. We have \begin{displaymath}
Reg(T) = \mathbb{E}\left(\sum_{l=1}^L \mathbb{E}\bigg(|\mathcal{E}_l| \Big(R(\mathbf{a}^*)-R\left(\mathbf{a}^l\right)\Big) ~|~ \mathcal{H}_{l-1}\bigg) \right)  = \mathbb{E}\left(\sum_{l=1}^L \left(1+ \sum_{k=1}^K \lambda_k \alpha_k^l \right)\left( R(\mathbf{a}^*)-R\left(\mathbf{a}^l\right)\right) \right).
\end{displaymath} To aid readability, we will define $\Delta R^l= \Big(1+ \sum_{k=1}^K \lambda_k \alpha_k^l \Big)\Big( R(\mathbf{a}^*)-R\left(\mathbf{a}^l\right)\Big)$ for each epoch $l \in [L]$ so that we have \begin{displaymath}
Reg(T) = \mathbb{E}\bigg( \sum_{l=1}^L \Delta R^l \bigg).
\end{displaymath} 

Next, we define a series of events $\mathcal{B}_l, l \in [L]$ which concern the value of the UCBs, as follows,
\begin{displaymath}
\mathcal{B}_l = \bigcup_{j=1}^J \Bigg\{\alpha_{j,l}^{UCB} \notin \bigg[\alpha_j, \alpha_j +  (1+\sqrt{2})\sqrt{\frac{8\alpha_j\log(Jl^2/2)}{\Lambda_{j,l}}} +  \frac{4(2+\sqrt{2})\log(Jl^2/2)}{\Lambda_{j,l}} \bigg] \Bigg\}.
\end{displaymath} We can bound the probability of this event using the concentration results derived in Lemma \ref{lemma::epochUCBconc}. Specifically, we have that \begin{align}
P(\mathcal{B}_l) &= \sum_{j=1}^J P(\alpha_{j,l}^{UCB} < \alpha_j) + P\bigg( \alpha_{j,l}^{UCB} > \alpha_j +  (1+\sqrt{2})\sqrt{\frac{8\alpha_j\log(Jl^2/2)}{\Lambda_{j,l}}} +  \frac{4(2+\sqrt{2})\log(Jl^2/2)}{\Lambda_{j,l}}\bigg) \nonumber \\
&\leq \sum_{j=1}^J \frac{3}{Jl} + P\bigg( |\bar\alpha_{j,l}^{UCB} - \alpha_j| >  (1+\sqrt{2})\sqrt{\frac{8\alpha_j\log(Jl^2/2)}{\Lambda_{j,l}}} +  \frac{4(2+\sqrt{2})\log(Jl^2/2)}{\Lambda_{j,l}}\bigg)  \label{eq::BlStepOne}
\end{align} since $P(\alpha_{j,l}^{UCB} < \alpha_j)$ is bounded for each $j \in [J]$ whether $\min(1,2\bar\alpha_{j,l})$ is $1$ or $2\bar\alpha_{j,l}$. Fixing $j \in [J]$ and considering a single summand from \eqref{eq::BlStepOne} we have, \begin{align}
&P\bigg( |\bar\alpha_{j,l}^{UCB}-\bar\alpha_{j}(l)|+|\bar\alpha_{j}(l)-\alpha_j| > (1+\sqrt{2})\sqrt{\frac{8\alpha_j\log(Jl^2/2)}{\Lambda_{j,l}}} +  \frac{4(2+\sqrt{2})\log(Jl^2/2)}{\Lambda_{j,l}}\bigg) \nonumber \\
&\leq P\bigg( |\bar\alpha_{j,l}^{UCB}-\bar\alpha_{j}(l)| >  \sqrt{\frac{16\alpha_j\log(Jl^2/2)}{\Lambda_{j,l}}}+ \frac{4(1+\sqrt{2})\log(Jl^2/2)}{\Lambda_{j,l}} \bigg) \nonumber \\
&\quad \quad   +P\bigg(|\bar\alpha_{j}(l)-\alpha_j| > \sqrt{\frac{8\alpha_j\log(Jl^2/2)}{\Lambda_{j,l}}}+ \frac{4\log(Jl^2/2)}{\Lambda_{j,l}}\bigg) \nonumber \\
&\leq P\bigg( \sqrt{\frac{4\min(1,2\bar\alpha_{j,l})\log(Jl^2/2)}{\Lambda_{j,l}}}+ \frac{4\log(Jl^2/2)}{\Lambda_{j,l}} >  \sqrt{\frac{16\alpha_j\log(Jl^2/2)}{\Lambda_{j,l}}}+ (1+\sqrt{2})\frac{4\log(Jl^2/2)}{\Lambda_{j,l}} \bigg) + \frac{4}{Jl} \nonumber  \\
&= P\bigg( \sqrt{\frac{4\min(1,2\bar\alpha_{j,l})\log(Jl^2/2)}{\Lambda_{j,l}}}  >  \sqrt{\frac{16\alpha_j\log(Jl^2/2)}{\Lambda_{j,l}}}+ \frac{4\sqrt{2}\log(Jl^2/2)}{\Lambda_{j,l}} \bigg) + \frac{4}{Jl} \nonumber \\
&\leq P\bigg( \frac{8\bar\alpha_{j,l}\log(Jl^2/2)}{\Lambda_{j,l}}  >  \frac{16\alpha_j\log(Jl^2/2)}{\Lambda_{j,l}}+ \frac{32\log^2(Jl^2/2)}{\Lambda_{j,l}^2} \bigg) + \frac{4}{Jl} \nonumber \\
&=  P\bigg( \bar\alpha_{j,l} >  2\alpha_j+ \frac{4\log(Jl^2/2)}{\Lambda_{j,l}} \bigg) + \frac{4}{Jl} \leq  \frac{6}{Jl}. \label{eq::BlStepTwo}
\end{align} The first inequality uses the triangle inequality, the second an application of equation \eqref{eq::UCBConcpart1}, the third bounds by replacing the minimum with the $2\bar\alpha_{j,l}$ term, and the final uses \eqref{eq::UCBconcpart3}. It follows, combining \eqref{eq::BlStepOne} and \eqref{eq::BlStepTwo}, that for any $l \in [L]$, $P(\mathcal{B}_l) \leq 9/l$.

Having established $\mathcal{B}_l$ as a low probabilitiy event we will decompose the regret according to $\mathcal{B}_l$, and bound it separately under the events $\mathcal{B}_l$ and $\neg\mathcal{B}_l$. Under $\mathcal{B}_l$ we will resort to trivial bounds on regret coming from the maximum of the reward function, but these will make limited contribution to the overall expected regret, since $\mathcal{B}_l$ occurs with low probability. On $\neg\mathcal{B}_l$, the parameters are bounded in a way that we can exploit to bound the per-round regret with quantities leading to an optimal overall bound. Specifically, we have for $l \in [L]$, \begin{align}
\mathbb{E}\big(\Delta R^l) &= \mathbb{E}\bigg( \Delta R^l \mathbb{I}\{\mathcal{B}_l\} + \Delta R^l \mathbb{I}\{\neg \mathcal{B}_l\}\bigg) \nonumber \\
&\leq (K+1)P(\mathcal{B}_l) + \mathbb{E}\big(\Delta R^l \mathbb{I}\{\neg \mathcal{B}_l\}\big) \nonumber \\
&\leq \frac{9(K+1)}{l} + \mathbb{E}\Bigg(\Big(1+ \sum_{k=1}^K \lambda_k \alpha_k^l \Big)\Big( R(\mathbf{a}^*)-R(\mathbf{a}^l)\Big)\mathbb{I}\{\neg\mathcal{B}_l\} \Bigg). \label{eq::decomp_by_Bl}
\end{align}

Since the reward function $R$ is monotonically increasing in the attractiveness parameter vector, and under $\neg \mathcal{B}_l$ we have ${\alpha}_{j,l}^{UCB} \geq \alpha_j$ for all $j \in [J]$, it follows that we also have \begin{displaymath}
R(\mathbf{a}, {\bs\alpha}_l^{UCB}) \geq R(\mathbf{a},\bs\alpha), \enspace \forall \mathbf{a} \in \mathcal{A},
\end{displaymath} under $\neg \mathcal{B}_l$. We also have by definition of the UCB algorithm that $R(\mathbf{a}^l,{\bs\alpha}_l^{UCB}) \geq R(\mathbf{a}^*,\bar{\bs\alpha}_l)$, and thus under $\neg \mathcal{B}_l$ we have \begin{align}
R(\mathbf{a}^*)-R(\mathbf{a}^l) &\leq R(\mathbf{a}^l,{\bs\alpha}_l^{UCB})-R(\mathbf{a}^l,\bs\alpha) \nonumber \\
&= \frac{\sum_{k=1}^K \lambda_k \alpha_{l,a^l_k}^{UCB}}{1+ \sum_{k=1}^K \lambda_k \alpha_{l,a^l_k}^{UCB}}- \frac{\sum_{k=1}^K \lambda_k \alpha_{a^l_k}}{1+ \sum_{k=1}^K \lambda_k \alpha_{a^l_k}} \nonumber \\
&\leq \frac{\sum_{k=1}^K \lambda_k (\alpha_{l,a^l_k}^{UCB}-\alpha_{a^l_k})}{1+ \sum_{k=1}^K \lambda_k \alpha_{l,a^l_k}^{UCB}} \nonumber \\
&\leq \frac{\sum_{k=1}^K \lambda_k (\alpha_{l,a^l_k}^{UCB}-\alpha_{a^l_k})}{1+ \sum_{k=1}^K \lambda_k \alpha_{a^l_k}} \label{eq::reg_under_Bl}
\end{align}

Thus, combining \eqref{eq::decomp_by_Bl} and \eqref{eq::reg_under_Bl} we have the following bound on per-epoch regret, \begin{displaymath}
\mathbb{E}\big(\Delta R^l\big) \leq \frac{9(K+1)}{l} + \mathbb{E}\left(\sum_{k=1}^K \lambda_k \left((1+\sqrt{2})\sqrt{\frac{8\alpha_{a_k^l}\log(Jl^2/2)}{\Lambda_{a^l_k,l}}}+\frac{4(2+\sqrt{2})\log(Jl^2/2)}{\Lambda_{a^l_k,l}}\right) \right).
\end{displaymath}
Aggregating over $L$ epochs, it follows that $Reg(T)$ is upper bounded by \begin{align}
&~\mathbb{E}\left(\sum_{l=1}^L \bigg[ \frac{9(K+1)}{l} + \sum_{k=1}^K \lambda_k \left((1+\sqrt{2})\sqrt{\frac{8\alpha_{a_k^l}\log(Jl^2/2)}{\Lambda_{a^l_k,l}}}+\frac{4(2+\sqrt{2})\log(Jl^2/2)}{\Lambda_{a^l_k,l}}\right) \bigg] \right) \nonumber \\
&\leq 9(K+1)(\log(T)+1) + \mathbb{E}\Bigg(\sum_{l=1}^L  \sum_{k=1}^K \bigg(\sqrt{\frac{48\alpha_{a_k^l}\lambda_k^2\log(Jl^2/2)}{\lambda_K\sum_{s=1}^l\mathbb{I}\{a_k^l \in \mathbf{a}_s\}}}+\frac{14\log(Jl^2/2)}{\lambda_K\sum_{s=1}^l\mathbb{I}\{a_k^l \in \mathbf{a}_s\}}\bigg)\Bigg) \nonumber \\
&\leq \left(9(K+1) + \frac{14KJ}{\lambda_K }\log(JT^2/2)\right)(\log(T)+1) + \nonumber \sum_{j=1}^J\mathbb{E}\Bigg( \sqrt{\frac{192}{\lambda_K}\alpha_j\lambda_1^2\log(Jl^2/2)\sum_{s=1}^L\mathbb{I}\{j \in \mathbf{a}_s\}}\Bigg) \nonumber \\
&\leq \left(9(K+1) + \frac{14JK}{\lambda_K }\log(JT^2/2)\right)(\log(T)+1) + \sqrt{\frac{192\lambda_1^2\log(Jl^2/2)J\mathbb{E}\left(\sum_{j=1}^J\sum_{s=1}^L\alpha_j\mathbb{I}\{a_k^l \in \mathbf{a}_s\}\right)}{\lambda_K}} \nonumber \\
&\leq \left(9(K+1) + \frac{14JK}{\lambda_K }\log(JT^2/2)\right)(\log(T)+1) + \sqrt{\frac{192\lambda_1^2\log(JT^2)JT}{\lambda_K}}. \enspace  \nonumber
\end{align} Here the second inequality replaces sums of reciprocals with increasing functions of $T$ (since $T>L$), the third inequality uses Cauchy-Schwarz, and the final inequality uses the property that $\sum_{j=1}^J\sum_{s=1}^L\alpha_j\mathbb{I}\{a_k^l \in \mathbf{a}_s\}$ is less than $T$ in expectation, as derived in \cite[Appendix A.3]{AgrawalEtAl2019}.  $\square$ 

\subsection{Proof of Proposition \ref{prop::UPBregret}}

The structure of the proof is similar to that of Proposition \ref{thm::PBregret}. We again decompose the regret as \begin{align*}
    Reg(T)=\mathbb{E}\left(\sum_{l=1}^L \Delta R^l\right) = \mathbb{E}\left(\sum_{l=1}^L \left(1+ \sum_{k=1}^K \lambda_k \alpha_k^l\right)\left(R(\mathbf{a}^*)-R\left(\mathbf{a}^l\right)\right) \right),
\end{align*} and define a series of low-probability events $\mathcal{C}_l$, $l\in [L]$, where the UCBs deviate substantially from their expected value, \begin{align*}
    \mathcal{C}_l = \bigcup_{j=1}^J \left\{ \alpha_{j,l}^{UCB-U} \notin \left[\alpha_j, \alpha_j + 2\sqrt{36\beta_{i,l,j}^2\log(Jl^2)}~ \right]  \right\}
\end{align*} 

For each $l\in[L]$, we may bound the probability of $\mathcal{C}_l$ as, \begin{align*}
    P\left(\mathcal{C}_l\right) &\leq \sum_{j=1}^J P\left(\alpha_{j,l}^{UCB-U}<\alpha_j\right) + P\left(\alpha_{j,l}^{UCB-U}> \alpha_j + 2\sqrt{36\beta_{i,l,j}^2\log(Jl^2)} ~\right) \leq \frac{4}{l^2}.
\end{align*} 

We now decompose the per-epoch regret on the low-probability event $\mathcal{C}_l$. For each $l\in [L]$ we have, \begin{align}
    \mathbb{E}\left(\Delta R^l\right) &\leq \mathbb{E}\left(\Delta R^l~|~\mathcal{C}_l\right)+ \mathbb{E}\left(\Delta R^l~|~\neg ~\mathcal{C}_l\right) \nonumber \\
    &\leq \frac{4K+4}{l^2} + \mathbb{E}\left( \left(1+ \sum_{k=1}^K \lambda_k \alpha_k^l\right)\left(R(\mathbf{a}^*)-R\left(\mathbf{a}^l\right)\right) ~\big| ~ \neg ~ \mathcal{C}_l\right). \label{eq::perepochUPB}
\end{align} The result \eqref{eq::reg_under_Bl} holds irrespective of the knowledge of position bias, so we have the following bound on per-epoch regret, combining \eqref{eq::reg_under_Bl}, \eqref{eq::perepochUPB}, and then utilising Lemma \ref{lem::betabound}, \begin{align*}
    \mathbb{E}\left(\Delta R^l\right) &\leq \frac{4K+4}{l^2} + \mathbb{E}\left(\sum_{k=1}^K 2\lambda_k\sqrt{36\beta_{1,a^l_k,l}^2\log(Jl^2)} \right)\\
    &\leq \frac{4K+4}{l^2} + \sum_{k=1}^K 2\lambda_k \mathbb{E}\left(\sqrt{\frac{36CJK\log(Jl^2)}{\max_{k'\in[K]}\tilde{N}_{k'a^l_k}}}\right).
\end{align*} Aggregating over $L$ epochs, we reach the stated regret bound as follows, \begin{align*}
    Reg(T) &\leq \mathbb{E}\left(\sum_{l=1}^L \frac{4K+4}{l^2} + \sum_{k=1}^K 2\lambda_k \sqrt{\frac{36CJK\log(Jl^2)}{\max_{k'\in[K]}\tilde{N}_{k'a^l_k}}} \right) \\
    &\leq 8(K+1) + \mathbb{E}\left( \sum_{l=1}^L\sum_{k=1}^K 2\lambda_k \sqrt{\frac{36CJK\log(Jl^2)}{\max_{k'\in[K]}\tilde{N}_{k'a^l_k}}} \right) \\
    &\leq 8(K+1) + \sqrt{144CJK^4T\log(JT^2)}. ~ \square
\end{align*}

\section{Proof of Regret Lower Bound} \label{app::lb}

\noindent \emph{Proof:} We first introduce some further notation. For each action $\mathbf{a} \in \mathcal{A}$, a fixed position bias vector $\bs\lambda$, and some constant $\epsilon \in (0,1/2]$ to be fixed later, define the attractiveness parameter vector $\bs\tau_\mathbf{a} \in (0,1]^{J+1}$ such that \begin{equation}
\tau_{\mathbf{a},j} = \begin{cases} &1, \quad \quad \quad \quad \quad j=0, \\
						&\frac{1}{S_K}+ \frac{\epsilon\lambda_k}{S_{K,2}}, \quad j = a_k,~ k \in [K], \\
						&\frac{1}{S_K}, \quad \quad \quad  \quad \text{ otherwise.}\end{cases}
\end{equation} Let $P_\mathbf{a}$ and $\mathbb{E}_\mathbf{a}$ denote the law and expectation with respect to the parametrisation $\alpha_j=\tau_{\mathbf{a},j}$. Under $P_{\mathbf{a}}$, the action $\mathbf{a}$ is optimal. 

We will also make use of the additional laws $P_{\mathbf{a} \setminus j'}$ and expectations $\mathbb{E}_{\mathbf{a} \setminus j'}$ for $j' \in \mathbf{a}$, for each $\mathbf{a} \in \mathcal{A}$. Under $P_{\mathbf{a} \setminus j'}$ we set $\alpha_j = \tau_j$, for all $j \neq j'$ and have $\alpha_{j'}=1/S_K$. We define $\mathcal{A}'$ as the set of incomplete orderings, i.e. orderings in $\mathcal{A}$ with a single unfilled slot. Each incomplete ordering $\mathbf{a}' \in \mathcal{A}$ has a corresponding law $P_{\mathbf{a}'}$ and expectation operator $\mathbb{E}_{\mathbf{a}'}$ which coincide with $P_{\mathbf{a}\j'}$ and $\mathbb{E}_{\mathbf{a}\j'}$ when placing $j'$ in the empty slot of $\mathbf{a}'$ would generate ordering $\mathbf{a}$.

Further, for $j \in \mathbf{a}$, introduce the notation $a^{-1}(j)$ to refer to the slot in which action $\mathbf{a}$ places item $j$ - i.e. $a_{a^{-1}(j)}=j$. For $j \notin \mathbf{a}$, $a^{-1}(j)=K+1$, and we let $\lambda_{K+1}=0$.

For a fixed $\mathbf{a} \in \mathcal{A}$, consider the per-round regret under the problem with parameters $\bs\tau_\mathbf{a}$. We have, for any $t \in [T]$, \begin{align*}
\mathbb{E}_{\mathbf{a}}(r(\mathbf{a})-r(\mathbf{a}_t)) &= \mathbb{E}_\mathbf{a}\left(\frac{1 + \epsilon}{2 + \epsilon} -  \frac{1+ \frac{\epsilon}{S_{K,2}} \sum_{k=1}^K \lambda_k \sum_{j \in \mathbf{a}} \lambda_{a^{-1}(j)} \mathbb{I}\{a_{k,t}=j\}}{2 + \frac{\epsilon}{S_{K,2}} \sum_{k=1}^K \lambda_k \sum_{j \in \mathbf{a}} \lambda_{a^{-1}(j)}  \mathbb{I}\{a_{k,t}=j\} }\right) \\
&= \mathbb{E}_\mathbf{a} \left( \frac{\epsilon - \frac{\epsilon}{S_{K,2}}\sum_{k=1}^K \lambda_k \sum_{j \in \mathbf{a}} \lambda_{a^{-1}(j)}  \mathbb{I}\{a_{k,t}=j\}}{(2+\epsilon)\left(2+ \frac{\epsilon}{S_{K,2}}\sum_{k=1}^K \lambda_k \sum_{j \in \mathbf{a}} \lambda_{a^{-1}(j)}  \mathbb{I}\{a_{k,t}=j\}\right)} \right) \\
&\geq \mathbb{E}_\mathbf{a} \left( \frac{\epsilon - \frac{\epsilon}{S_{K,2}}\sum_{k=1}^K \lambda_k \sum_{j \in \mathbf{a}} \lambda_{a^{-1}(j)} \mathbb{I}\{a_{k,t}=j\}}{7} \right). 
\end{align*} 
\noindent It follows that in $T$ rounds, the regret satisfies \begin{align}
\max_{\bs\alpha \in (0,1]^{J}} Reg_{\bs\alpha}(T) &\geq \max_{\mathbf{a} \in \mathcal{A}} \mathbb{E}_{\mathbf{a}} \left(\sum_{t=1}^T r(\mathbf{a})-r(\mathbf{a}_t)\right) \nonumber \\
&\geq \frac{1}{|\mathcal{A}|}\sum_{\mathbf{a} \in \mathcal{A}} \mathbb{E}_{\mathbf{a}} \left(\sum_{t=1}^T r(\mathbf{a})-r(\mathbf{a}_t)\right) \nonumber \\
&\geq \frac{1}{7|\mathcal{A}|}\sum_{\mathbf{a} \in \mathcal{A}}\mathbb{E}_\mathbf{a}\left( \epsilon T - \frac{\epsilon}{S_{K,2}} \sum_{j \in \mathbf{a}} \lambda_{a^{-1}(j)}\sum_{t=1}^T\lambda_{a_t^{-1}(j)}\right). \label{eq::regboundinitial}
\end{align} Modifying slightly the notation from Section \ref{sec::algorithms}, we consider the innermost sum $\Lambda_j(T)=\sum_{t=1}^T\lambda_{a_t^{-1}(j)}$ in isolation, for $j\in[J]$.


For an $\mathbf{a} \in\mathcal{A}$, and $j \in \mathbf{a}$ the expectation of these random variables has the following property, \begin{align*}
\mathbb{E}_\mathbf{a}(\Lambda_{j}(T)) &\leq \mathbb{E}_{\mathbf{a} \setminus j}(\Lambda_{j}(T)) + \left|\mathbb{E}_{\mathbf{a} \setminus j}(\Lambda_{j}(T))-\mathbb{E}_\mathbf{a}(\Lambda_{j}(T))\right| \\
&\leq \mathbb{E}_{\mathbf{a} \setminus j}(N_{kj}(T)) + T \lambda_{max}||P_{\mathbf{a} \setminus j}-P_\mathbf{a}||_{TV} \\
&\leq  \mathbb{E}_{\mathbf{a} \setminus j}(\Lambda_{j}(T)) + T \lambda_{max}\sqrt{\frac{KL(P_{\mathbf{a} }~||~P_{\mathbf{a}\setminus j})}{2}}.
\end{align*} Here, the second inequality uses the discrete support of $\Lambda_j(T)$, and the final inequality uses Pinsker's inequality.

We now turn our attention to the KL divergence term $KL(P_{\mathbf{a}}~||~P_{\mathbf{a}\setminus j})$. By the Law of Total Entropy (see e.g. Theorem 2.5.3 of \cite{CoverThomas2012}), we have \begin{align}
KL(P_{\mathbf{a}}~||~P_{\mathbf{a}\setminus j}) &= \sum_{t=1}^T KL\left(P_\mathbf{a}(Q_t ~|~ Q_1,\dots,Q_{t-1} ) ~||~ P_{\mathbf{a} \setminus j'}(Q_t ~|~ Q_1,\dots,Q_{t-1})\right) \nonumber \\
&= \sum_{t=1}^T\sum_{\mathbf{a}' \in \mathcal{A}: j \in  \mathbf{a}'} \mathbb{P}_{\mathbf{a'}}\{\mathbf{a}_t = \mathbf{a}'\} KL(P_\mathbf{a}(Q ~|~ \mathbf{a}') ~||~ P_{\mathbf{a} \setminus j'}(Q ~|~ \mathbf{a}')) \nonumber \\
&\leq \sum_{k=1}^K  \max_{\mathbf{a}' \in \mathcal{A}:a_k'=j} \mathbb{E}_{\mathbf{a}'}(N_{kj}(T))KL(P_{\mathbf{a}}(Q ~|~ \mathbf{a}') ~||~ P_{\mathbf{a} \setminus j}(Q ~|~ \mathbf{a}')) \label{eq::KLdecomposed}
\end{align} Here we also use the property that any algorithm gives a deterministic mapping from $Q_{1:t-1}$ to $\mathbf{a}_t$ - since even an instance of a `randomised' algorithm can, alternatively, be viewed as a deterministic algorithm randomly selected from a (potentially infinitely large) population of algorithms.

 The following lemma, whose proof is given in the the following subsection, bounds the contribution to the KL divergence from a single round.

\begin{lemma} \label{lem::KLbound}
For two actions $\mathbf{a},\mathbf{a}' \in \mathcal{A}$, and an item $j \in \mathbf{a}$  such that $a_n=j$ and $a_{n'}' =j$ for some (possibly equal) $n,n' \in [K]$, we have the following bound on the KL divergence between $P_{\mathbf{a}}(Q ~|~ \mathbf{a}')$ and $P_{\mathbf{a} \setminus j'}(Q ~|~ \mathbf{a}')$, the marginal distributions on $Q~|~\mathbf{a}'$ implied by the laws $P_\mathbf{a}$ and $P_{\mathbf{a} \setminus j}$, \begin{equation}
KL\left(P_\mathbf{a}(Q ~|~ \mathbf{a}') ~||~ P_{\mathbf{a}\setminus j}(Q ~|~ \mathbf{a}')\right) \leq \frac{17\epsilon^2\lambda_{{a'}^{-1}(j)}^2S_K}{S_{K,2}^2}. \label{eq::lemKLboundstatement}
\end{equation}
\end{lemma} 

Thus, combining the decomposition of KL divergence in \eqref{eq::KLdecomposed} and the bound on per-round KL divergence in \eqref{eq::lemKLboundstatement} we have for any $j \in \mathbf{a}$, \begin{displaymath}
KL(P_{\mathbf{a}} ~||~ P_{\mathbf{a} \setminus j}) \leq \sum_{k=1}^K \mathbb{E}_{\mathbf{a}}\left(N_{kj}(T)\right) \frac{17 \epsilon^2 \lambda_{k}^2S_K}{S_{K,2}^2} 
\end{displaymath} 
Then combining with \eqref{eq::regboundinitial} we have that the regret is lower bounded, similarly to in \cite{ChenWang2018}, as, \begin{align*}
Reg(T) &\geq \frac{1}{7|\mathcal{A}|} \sum_{\mathbf{a} \in \mathcal{A}}  \left[\epsilon T - \frac{\epsilon}{S_{K,2}} \sum_{j\in\mathbf{a}} \lambda_{a^{-1}(j)}\Lambda_j(T) \right] \\
&\stackrel{(a)}{\geq} \frac{\epsilon T}{7} - \frac{1}{7|\mathcal{A}|}\sum_{\mathbf{a} \in \mathcal{A}} \left[ \frac{\epsilon}{S_{K,2}} \sum_{j \in \mathbf{a}} \lambda_{a^{-1}(j)} \left(\mathbb{E}_{\mathbf{a} \setminus j}(\Lambda_j(T)) + T\lambda_{max}\sqrt{\frac{KL(P_{\mathbf{a}}~||~P_{\mathbf{a}\setminus j})}{2}}~\right)\right]  \\
&\stackrel{(b)}{\geq} \frac{\epsilon T}{7} - \frac{\epsilon}{7S_{K,2}|\mathcal{A}|}\sum_{j=1}^J \sum_{\mathbf{a}\in\mathcal{A}: j\in\mathbf{a}}\left[\lambda_{a^{-1}(j)} \left(\mathbb{E}_{\mathbf{a} \setminus j}(\Lambda_j(T)) + T\lambda_{max}\sqrt{\frac{KL(P_{\mathbf{a}}~||~P_{\mathbf{a}\setminus j})}{2}}~\right)\right]\\
&= \frac{\epsilon T}{7} - \frac{\epsilon}{7S_{K,2}|\mathcal{A}|}\sum_{\mathbf{a}'\in\mathcal{A}'}\sum_{j\not\in\mathbf{a}'} \left[\lambda_{\mathbf{a}'}\left(\mathbb{E}_{\mathbf{a}'}(\Lambda_j(T)) + T\lambda_{max}\sqrt{\frac{KL(P_{\mathbf{a}}~||~P_{\mathbf{a}\setminus j'})}{2}}~\right)\right] \\
&\stackrel{(c)}{\geq} \frac{\epsilon T}{7} - \frac{\epsilon}{7S_{K,2}|\mathcal{A}|}\sum_{\mathbf{a}'\in\mathcal{A}'}\lambda_{\mathbf{a}'}S_KT  - \frac{\epsilon |\mathcal{A}'|T\lambda_{max}}{7S_{K,2}|\mathcal{A}|}\max_{\mathbf{a}'\in\mathcal{A}'}\sum_{j'\not\in\mathbf{a}'}\sqrt{\frac{KL(P_{\mathbf{a}}~||~P_{\mathbf{a}\setminus j'})}{2}} \\
&\stackrel{(d)}{\geq} \frac{\epsilon T}{7} - \frac{\epsilon S_K K T}{7S_{K,2}(J-K+1)} - \frac{\epsilon KT}{7S_{K,2}}\sqrt{\max_{\mathbf{a}'\in\mathcal{A}'}\sum_{j\not\in\mathbf{a}'}\frac{\sum_{k=1}^K \mathbb{E}_{\mathbf{a}}\left(N_{kj'}(T)\right)17\epsilon^2\lambda^2_{\mathbf{a}'}S_K}
{2S_{K,2}^2(J+K-1)}} \\
&\stackrel{(e)}{\geq} \frac{\epsilon T}{7} - \frac{\epsilon C S_K T}{7(J-K+1)} -\sqrt{\frac{17CS_KT^3\epsilon^4}{2S_{K,2}(J-K+1)}}.
\end{align*}
Here inequality (a) introduces the bound on the $\Lambda_j$ terms, and inequality (b) reorders the summations. Inequality (c) replaces the $|\mathcal{A}'|$ individual summands with $|\mathcal{A}'|$ copies of their largest and notes any sum of $\Lambda_j(T)$ terms is necessarily bounded by $S_KT$. Then, inequality (d) uses the concavity of the square root, while noting $|\mathcal{A}'|/|\mathcal{A}|=K/(J-K+1)$. The final inequality introduces a constant $C>0$ such that $K/S_{K,2} <C$. 
Finally, we complete the proof by choosing $\epsilon=O(\sqrt{JS_{K,2}}/\sqrt{TS_K})$ and using the assumption that $K \leq J/4$. $\square$

\subsection{Proof of Lemma \ref{lem::KLbound}}
In this section we provide a proof of Lemma \ref{lem::KLbound}, which helps to complete the proof of the regret lower bound, Proposition \ref{thm::lb}. Lemma \ref{lem::KLbound} gives a bound on the KL divergence between the marginal distributions over a single click variable, under (particular) different attractiveness parameter vectors. The proof makes use of the following result, originally given as Lemma 3 in \cite{ChenWang2018}, bounding the KL-divergence between categorical random variables. 

\begin{lemma}[Lemma 3 of \cite{ChenWang2018}] \label{lem::CWlemma}
Suppose $P$ is a categorical distribution on $[M]_0$ with parameters $p_0,\dots,p_M$, such that if $X \sim P$, $P(X=m)=p_m$ for $m \in [M]_0$. Suppose also that $Q$ is an equivalently defined categorical distribution with parameters $q_0,\dots,q_M$, and we have $\delta_m=p_m-q_m$ for $m \in [M]_0$. Then \begin{equation*}
KL(P ~||~ Q) \leq \sum_{m=0}^M \frac{\delta_m^2}{q_m}.
\end{equation*} 
\end{lemma} 


\noindent \emph{Proof of Lemma \ref{lem::KLbound}:}  We begin by deriving expressions for the parameters $p_k := P_{\mathbf{a}}(Q=k ~|~ \mathbf{a}')$, \begin{align*}
p_0 &= \frac{1}{2+\epsilon\sum_{l=1}^K \lambda_l \sum_{m=1}^K \frac{\lambda_m}{S_{K,2}}\mathbb{I}\{a_l' =a_m \}}, \\
p_k &= \frac{\frac{\lambda_k}{S_K} + \epsilon\lambda_k \sum_{m=1}^K \frac{\lambda_m}{S_{K,2}}\mathbb{I}\{a_k' =a_m\} }{2+\epsilon\sum_{l=1}^K \lambda_l \sum_{m=1}^K \frac{\lambda_m}{S_{K,2}}\mathbb{I}\{a_l' =a_m\}}, \enspace k \in [K],
\end{align*} and $q_k:= P_{\mathbf{a} \setminus j}(Q=k ~|~ \mathbf{a}')$, \begin{align*}
q_0 &= \frac{1}{2 + \epsilon\sum_{l=1}^K \lambda_l \sum_{m=1}^K \frac{\lambda_m}{S_{K,2}}\mathbb{I}\{a_l' = a_m, \enspace m \neq n\}}, \\
q_k &=  \frac{\frac{\lambda_k}{S_K}+ \epsilon\lambda_k \sum_{m=1}^K \frac{\lambda_m}{S_{K,2}}\mathbb{I}\{a_k' = a_m, \enspace m \neq n\}}{2+ \epsilon\sum_{l=1}^K \lambda_l \sum_{m=1}^K \frac{\lambda_m}{S_{K,2}}\mathbb{I}\{a_l'=a_m, \enspace m \neq n\}}, \enspace k \in [K].
\end{align*} To apply Lemma \ref{lem::CWlemma} we consider the differences in these parameters. For the no-click event we have  \begin{align*}
p_0 - q_0 &=\frac{-\epsilon\lambda_n\lambda_{n'}}{S_{K,2}\left( 2+ \epsilon\sum_{l=1}^K \lambda_l \sum_{m=1}^K \frac{\lambda_m}{S_{K,2}}\mathbb{I}\{a_l'=a_m\}\right) \left( 2 + \epsilon\sum_{l=1}^K \lambda_l \sum_{m=1}^K \frac{\lambda_m}{S_{K,2}}\mathbb{I}\{a_l'=a_m, \enspace m \neq n\}\right)}.
\end{align*} For $k \in [K]$ such that $k \neq n'$ we have \begin{align*}
p_k-q_k &= \frac{\frac{\lambda_k}{S_K} + \epsilon\lambda_k \sum_{m=1}^K \frac{\lambda_m}{S_{K,2}}\mathbb{I}\{a_k' =a_m\} }{2+\epsilon\sum_{l=1}^K \lambda_l \sum_{m=1}^K \frac{\lambda_m}{S_{K,2}}\mathbb{I}\{a_l' =a_m\}}-  \frac{\frac{\lambda_k}{S_K}+ \epsilon\lambda_k \sum_{m=1}^K \frac{\lambda_m}{S_{K,2}}\mathbb{I}\{a_k' = a_m, \enspace m \neq n\}}{2+ \epsilon\sum_{l=1}^K \lambda_l \sum_{m=1}^K \frac{\lambda_m}{S_{K,2}}\mathbb{I}\{a_l'=a_m, \enspace m \neq n\}},\\
&= \frac{-\epsilon\frac{\lambda_k\lambda_{n}\lambda_{n'}}{S_KS_{K,2}} - \epsilon^2\frac{\lambda_k^2\lambda_{n}\lambda_{n'}}{S_{K,2}^2}}{\left( 2+\epsilon\sum_{l=1}^K \lambda_l \sum_{m=1}^K \frac{\lambda_j}{S_{K,2}}\mathbb{I}\{a_l'=a_m\}\right) \left( 2 + \epsilon\sum_{l=1}^K \lambda_l \sum_{m=1}^K \frac{\lambda_m}{S_{K,2}}\mathbb{I}\{a_l'=a_m, \enspace m \neq n\}\right)}.
\end{align*} Finally for the slot $n'$ in which item $j$ is placed we have \begin{align*}
p_{n'}-q_{n'} &=  \frac{\frac{\lambda_{n'}}{S_K} + \epsilon\frac{\lambda_{n}\lambda_{n'}}{S_{K,2}}}{2+\epsilon\sum_{l=1}^K \lambda_l \sum_{m=1}^K \frac{\lambda_m}{S_{K,2}}\mathbb{I}\{a_l' =a_m\}}- \frac{\frac{\lambda_{n'}}{S_K}}{2+\epsilon\sum_{l=1}^K \lambda_l \sum_{m=1}^K \frac{\lambda_m}{S_{K,2}}\mathbb{I}\{a_l' =a_m, m \neq n\}}\\
&=\frac{\frac{\lambda_n \lambda_{n'}}{S_{K,2}}\left(2\epsilon+\epsilon^2\sum_{l=1}^K \sum_{m=1}^K \frac{\lambda_l \lambda_m}{S_{K,2}}\mathbb{I}\{a_l'=a_m,m\neq n\} \right) - \epsilon\frac{\lambda_n\lambda_{n'}^2}{S_K S_{K,2}}}{\left( 2+\epsilon\sum_{l=1}^K \lambda_l \sum_{m=1}^K \frac{\lambda_m}{S_{K,2}}\mathbb{I}\{a_l'=a_m\}\right) \left( 2 + \epsilon\sum_{l=1}^K \lambda_l \sum_{m=1}^K \frac{\lambda_m}{S_{K,2}}\mathbb{I}\{a_l'=a_m, \enspace m \neq n\}\right)}.
\end{align*} 

It follows, subsequent to some further algebra, that we have \begin{align*}
\frac{(p_0-q_0)^2}{q_0} \leq \frac{\epsilon^2 \lambda_{n}^2\lambda_{n'}^2}{8S_{K,2}^2}, ~\frac{(p_k-q_k)^2}{q_k} \leq \frac{7\epsilon^2\lambda_k\lambda_{n}\lambda_{n'}^2}{8S_{K,2}^3/S_K}, \text{ and } \frac{(p_{n'}-q_{n'})^2}{q_{n'}} \leq \frac{9\epsilon^2\lambda_{n}\lambda_{n'}^2}{8S_{K,2}^2/S_K}.
\end{align*} As such, we have the following by Lemma \ref{lem::CWlemma}, and the assumption that $S_{K,2}\geq 1$ \begin{align*}
KL(P_{\mathbf{a}}(Q ~|~ \mathbf{a}') ~||~ P_{\mathbf{a} \setminus j}(Q ~|~ \mathbf{a}')) &\leq \frac{\epsilon^2\lambda_{n}^2\lambda_{n'}^2}{8S_{K,2}^2} + \sum_{k=1}^K \frac{7\epsilon^2\lambda_k\lambda_{n}\lambda_{n'}^2}{8S_{K,2}^3/S_K}\mathbb{I}\{k \neq n'\} + \frac{9\epsilon^2\lambda_{n}\lambda_{n'}^2}{8S_{K,2}^2/S_K} \\
&\leq \frac{17\epsilon^2\lambda_{n'}^2S_K}{S_{K,2}^2}.~ ~ ~ \square
\end{align*}

\section{Proofs of Technical Lemmas} \label{app::technical_proofs}

In this section we provide proofs of the remaining technical lemmas arising in the main text.

\subsection{Proof of Lemma \ref{lem::Geometric}} The probability of a no-click event given action $\mathbf{a}^l$ is given as \begin{displaymath}
p_0(\mathbf{a}^l) = P(Q_t=0|\mathbf{a}_t=\mathbf{a}^l) =  \frac{1}{1+\sum_{k=1}^K \lambda_k\alpha_k^l}.
\end{displaymath} It follows that $n^l = |\mathcal{E}_l|-1$, the number of clicks before the no-click event in epoch $l$ is a Geometric random variable with parameter $p_0(\mathbf{a}^l)$. It follows, that conditioned on $n^l$, each $n^l_{k}$ count may be viewed as a Binomial random variable,  \begin{displaymath}
n^l_{k} | n^l  \sim Binom(n^l, p_{k}),
\end{displaymath} where 
\begin{displaymath}
\tilde{p}_{k} = \frac{\lambda_k\alpha_k}{\sum_{v=1}^K \lambda_v\alpha_v},
\end{displaymath} is the probability of a click on the item in position $k$, given that there is a click.

The moment generating function of a Binomial random variable is of course well-known, and we therefore have \begin{displaymath}
\mathbb{E}_\pi(e^{\theta n_{k}^l}) = \mathbb{E}_{n^l}\big( \mathbb{E}_\pi(e^{\theta n_{k}^l} | n^l)\big) = \mathbb{E}_{n^l}\big( (\tilde{p}_{k}e^{\theta}+1-\tilde{p}_{k})^{n^l}\big).
\end{displaymath} We then consider the result that if $X$ is a Geometric random variable with parameter $p$, then for any $\tau$ such that $\tau(1-p)<1$  we have $\mathbb{E}(\tau^X)=p/(1-\tau(1-p))$. It follows that, for any $\theta<\log(\frac{\lambda_k\alpha_k^l+1}{\lambda_k\alpha_k^l})$, we have \begin{align*}
\mathbb{E}_\pi(e^{\theta n_{k}^l}) &= \frac{p_0}{1-(\tilde{p}_{k}e^{\theta/\lambda_k}+1-\tilde{p}_{k})(1-p_0) } \\
&= \frac{p_0}{1-\big(\tilde{p}_k(e^{\theta}-1)+1\big)(1-p_0)} \\
&=\frac{p_0}{1 - (1-p_0) - \lambda_k\alpha_k^lp_0(e^{\theta}-1)} \\
&= \frac{1}{1- \lambda_k\alpha_k^l (e^{\theta}-1)}, 
\end{align*} as stated.  We recognise that each $n_k^l ~|~ \mathbf{a}^l$ is an independent geometric random variable, by considering the moment generating function of the geometric random variable $X$ with parameter $p$ and density $f_X(k)=p(1-p)^k$ for $k \in \{0,1,2,\dots\}$, is given as \begin{displaymath}
M(t) = \frac{p}{1-e^t(1-p)} = \frac{1}{1-(\frac{1-p}{p})(e^t-1)},
\end{displaymath} and is defined only for $e^t(1-p)<1$. $\square$

\subsection{Proof of Lemma \ref{lem::EMconv}} We will first demonstrate that the log-likelihood function has at most one stationary point on the parameter space $(0,1]^{J+K-1}$, We have that the log-likelihood of data $\mathbf{N}$ given $\tilde{\mathbf{N}}$ and parameters $\bs\alpha, \bs\lambda$ is \begin{equation}
\log\mathcal{L}(\mathbf{N};\tilde{\mathbf{N}},\bs\alpha,\bs\lambda) = \sum_{k=1}^K \sum_{j=1}^J N_{kj}\log(\alpha_j\lambda_k) - (N_{kj}+\tilde{N}_{kj})\log(1+\alpha_j\lambda_k). \label{eq:llike}
\end{equation} Consdier the partial derivatives of the log-likelihood,  \begin{align*}
\frac{\partial \log\mathcal{L}}{\partial \alpha_j} &= \sum_{k=1}^K \frac{N_{kj}-\alpha_j\lambda_k \tilde{N}_{kj}}{\alpha_j(1+\alpha_j\lambda_k)}, \quad j \in [J], \quad \frac{\partial \log\mathcal{L}}{\partial \lambda_k} = \sum_{j=1}^J \frac{N_{kj}-\alpha_j\lambda_k \tilde{N}_{kj}}{\lambda_k(1+\alpha_j\lambda_k)}, \quad k \in [K] \setminus \{1\}.
\end{align*} The solutions of the system of equations $\partial \log\mathcal{L}/\partial \alpha_j =0, \partial\log\mathcal{L} /\partial\lambda_k=0$, $j \in [J]$, $k \in [K]\setminus\{1\}$, coincide with the solutions of, \begin{align}
\sum_{k=1}^K (N_{kj}-\alpha_j \tilde{N}_{kj})\prod_{m \neq k}(1+\alpha_j\lambda_m) &= 0, \quad j \in [J] \label{eq::rootsj} \\
\sum_{j=1}^J(N_{kj}-\lambda_k \tilde{N}_{kj})\prod_{i \neq j}(1+\alpha_i\lambda_k)&=0, \quad k \in [K] \setminus \{1\}. \label{eq::rootsk}
\end{align} We will demonstrate that the log-likelihood has at most one stationary point on $(0,1]^{J+K-1}$ by showing that the system of equations given by \eqref{eq::rootsj} and \eqref{eq::rootsk} has at most one solution on $(0,1]^{J+K-1}$.

For each $j \in [J]$ consider the LHS of equation \eqref{eq::rootsj} with all $\lambda_2,\dots,\lambda_K$ fixed in $(0,1]$. The result is an $O(\alpha_j^K)$ polynomial, which can be decomposed in to the summation of $k$ $O(\alpha_j^K)$ polynomials, $g_k(\alpha_j)=
(N_{kj}-\alpha_j \tilde{N}_{kj})\prod_{m \neq k}(1+\alpha_j\lambda_m)$, $k \in [K]$. We have roots of $g_k$ at  $\alpha_j=N_{kj}/\tilde{N}_{kj}$ and $\alpha_j=-1/\lambda_m$, $\forall m \neq k$, and notice that $g_k(\alpha_j)$ has negative leading order term when $\alpha_j>0$. Notice that only one of the solutions is positive, and lies in $(0,1]$ iff $0 < N_{kj} \leq \tilde{N}_{kj}$. 

Since each polynomial $g_k$ has negative leading order term, it follows that $\sum_{k=1}^K g_k(\alpha_j) =0$, i.e. equation \eqref{eq::rootsj}, also has at most one solution in $(0,1]$, for fixed $\lambda_2,\dots,\lambda_K$. An analogous argument applied to equation \eqref{eq::rootsk} tells us that there is at most one positive solution value in $(0,1]$ for each $\lambda_k$, $k \in [K] \setminus \{1\}$, coinciding with all other variables being positive.

This tells us that the system of equations where the partial derivatives are set to zero, has at most one solution in $(0,1]^{K+J-1}$ and as such the log-likelihood function has at most one stationary point on $(0,1]^{K+J-1}$. From this we deduce that the log-likelihood is either monotonic on $(0,1]^{K+J-1}$ or unimodal. We have from \cite{Wu1983} that the EM algorithm will converge to the unique MLE if the log-likelihood is unimodal and continuous, and thus that the EM algorithm, Algorithm \ref{alg::EMinf}, will converge to the MLEs. $\square$

\section{Independent Product Parameter Model} \label{app::suboptimal}

A perhaps more straightforward approach to the unknown position bias variant of the MNL-LTR problem would be to exploit the closed-form distribution of the estimators $\bar\gamma_{j,k}(L)=N_{kj}/\tilde{N}_{kj}$ of product parameters $\gamma_{j,k} = \alpha_j\lambda_k$, $j \in [J]$, $k \in [K]$ and build UCBs around these parameters independently. In this section we argue that this is not an appropriate strategy. Specifically, although such an approach can be shown to eventually learn the optimal action, and indeed have sublinear regret, the amount of exploration it requires is prohibitively large in comparison to our proposed strategy.

\subsection{An MNL-Bandit Approach to MNL-LTR with Unknown Position Biases}

We notice that by modelling the $\gamma_{jk}$ parameters as independent, the unknown position bias variant of MNL-LTR can also be thought of as a \emph{constrained MNL-bandit} problem. We can design such a formulation where the decision-maker is oblivious to the ranking aspect, but the constraints on their actions ensure that they implicitly assign a single item to a single slot. 

In such a setting there are $JK$ \emph{objects}, indexed $S_{j,k}$ for $j\in[J]$, $k\in[K]$, where selecting object $S_{j,k}$ represents placing item $j$ in slot $k$. Object $S_{j,k}$ is therefore associated with parameter $\gamma_{jk}$. In each round $t \in [T]$ the decision-maker chooses a set $\mathbf{S}_t$ containing $K$ of the objects. The constraints of the associated MNL-LTR problem are such that in each round exactly one of these objects must have index $k$ for each $k \in [K]$. 

Introducing $JK$ further indicator variables $x_{jk}(t)=\mathbb{I}\{S_{j,k}\in\mathbf{S}_t\}$ for each $t \in [T]$, these constraints may be expressed as, \begin{equation} 
\sum_{j=1}^J x_{j,k}(t)=1 ~\forall k \in [K], \quad \text{and} \quad \sum_{k=1}^K x_{j,k}(t) \leq 1 ~ \forall j \in [J], \quad \forall t \in [T]. \label{eq::constraints}
\end{equation} The first constraint captures the rule that every slot is utilised, and the second constraint captures the rule that each item is used at most once. A valid set of objects $\mathbf{S}_t$, satisfying \eqref{eq::constraints}, maps in a one-to-one fashion to a valid action $\mathbf{a}_t \in \mathcal{A}$ for our MNL-LTR problem.

Both \cite{AgrawalEtAl2017} and \cite{AgrawalEtAl2019} derive order-optimal guarantees for constrained MNL-bandit algorithms. However, the constraint considered in \cite{AgrawalEtAl2017} is only a simple cardinality constraint - i.e. an upper limit on the number of objects chosen in each round. Thus the guarantees on the Thompson Sampling approach proposed therein do not carry to the problem with constraints \eqref{eq::constraints}. \cite{AgrawalEtAl2019}, however, allow for a more general class of constraints, requiring that constraints may be expressed in the form $A\mathbf{x} \leq \mathbf{b}$, where $A$ is a totally unimodular (TU) matrix, and $\mathbf{b}$ is an integer-valued vector. The $JK \times (J+2K)$ matrix $A$ implied by constraints \eqref{eq::constraints} can be shown to be TU. Thus the regret guarantees on the UCB algorithm proposed in \cite{AgrawalEtAl2019} apply directly to the constrained MNL-bandit instance associated with an MNL-LTR instance. 

Algorithm \ref{alg::MNLB} describes a modification of such an algorithm to incorporate our sharper concentration results. Then, the corollary below gives the corresponding order result on regret enjoyed by this algorithm. The proof of this result is omitted as it follows directly from the observation that the coefficient matrix implied by constraints \eqref{eq::constraints} is TU, and substituting the concentration results of Lemma \ref{lemma::epochUCBconc} in to the proof of Theorem 1 of \cite{AgrawalEtAl2019}.

\begin{corollary}
There exist constants $C_1,C_2>0$ such that for any MNL-LTR problem where the item attractiveness parameters satisfy $\alpha_j \leq \alpha_0 =1$, $j \in[J]$ and the position biases satisfy $\lambda_1=1$, $\lambda_k \leq \lambda_1$ $k \geq 2$, the regret in $T$ rounds of Algorithm \ref{alg::MNLB} satisfies \begin{displaymath}
Reg(T) \leq C_1\sqrt{JKT\log(JKT^2) } + C_2JK\log^2(JKT).
\end{displaymath}
\end{corollary}

Although this implies a guarantee on the performance of Algorithm \ref{alg::MNLB} which is near optimal in its dependence in $T$, we see that the $O(\log(T))$ term has a worse dependence on $JK$ than, for instance, the Epoch-UCB algorithm for the known position bias case. This hints to the critical issue with deploying Algorithm \ref{alg::MNLB}, that its exploration cost scales linearly with the \emph{product} of the number of items and number of slots. In the experiments in Section \ref{sec::experiments}, in particular problem (c), we see that even in 'simple' problems - where the optimal action can be identified quickly - Algorithm \ref{alg::MNLB} is slow to converge.

\begin{algorithm}[]
    \caption{Epoch-UCB algorithm for MNL-bandit model of MNL-LTR}
    \label{alg::MNLB}
    \vspace{0.2cm}    
     Initialise with $l=0$, and $Q_0=0$. Iteratively perform the following for $t \in [T]$,  \vspace{0.2cm}
     
     If $Q_{t-1}=0$ \begin{itemize}
     \item Set $l \leftarrow l+1$
     \item Calculate UCBs. For $j \in [J]$, $k \in [K]$ compute,  $$\gamma_{j,k,l}^{UCB}= \bar\gamma_{j,k}(l-1)+ \sqrt{\frac{4\min(1,2\bar\gamma_{j,k}(l-1))\log(JKl^2/2)}{\sum_{s=1}^{l-1}\mathbb{I}\{S_{j,k}\in\mathbf{S}_l\}}} + \frac{{4\log(JKl^2/2)}}{\sum_{s=1}^{l-1}\mathbb{I}\{S_{j,k}\in\mathbf{S}_l\}}.$$
     \item Solve the optimisation problem for object indicator variables \begin{align*}
     \mathbf{x}_l=\argmax_{x_{j,k}, j\in [J], k\in[K]} \sum_{j=1}^J \sum_{k=1}^K &x_{j,k}\gamma_{j,k,l}^{UCB} \\
     \text{s.t. } \sum_{j=1}^J x_{j,k}(t)&=1 ~\forall k \in [K],  \\
     \quad \quad \sum_{k=1}^K x_{j,k}(t) &\leq 1 ~ \forall j \in [J]
     \end{align*}
     \item Select an action $\mathbf{a}_t \in \mathcal{A}$ associated with $\mathbf{x}_l \in \{0,1\}^{J\times K}$, and observe click variable $Q_t$,
     \end{itemize}
     otherwise, set action $\mathbf{a}_t=\mathbf{a}_{t-1}$, and observe click variable $Q_t$.
        \vspace{0.2cm}
\end{algorithm}

\end{document}